\documentclass{article}

\usepackage[main, final]{neurips_2026}

\usepackage[T1]{fontenc}    
\usepackage{nicefrac}       
\usepackage{microtype}      

\usepackage{float}
\usepackage[dvipsnames]{xcolor}
\usepackage{array}
\usepackage{colortbl}
\usepackage{graphicx}
\usepackage{epigraph}
\usepackage{subcaption}
\usepackage[autolanguage]{numprint}
\usepackage[utf8]{inputenc}
\usepackage{csquotes}
\usepackage{soul}
\usepackage{amssymb}
\usepackage{booktabs}
\usepackage{amsmath}
\usepackage{multirow}
\usepackage{siunitx}
\usepackage{xcolor}
\usepackage{graphicx} 
\usepackage{algorithm}
\usepackage{algpseudocode}
\usepackage{amsmath}
\usepackage{listings}
\usepackage{overpic}
\usepackage{dashrule}
\usepackage{enumitem} 
\newcommand{\best}{\cellcolor{tealdark}}
\newcommand{\sbest}{\cellcolor{teallight}}

\usepackage{tikz}
\definecolor{cvprblue}{rgb}{0.21,0.49,0.74}
\definecolor{yellow}{rgb}{1, 1, 0.7}
\definecolor{orange}{rgb}{1, 0.85, 0.7}
\definecolor{tablered}{rgb}{1, 0.7, 0.7}
\definecolor{tealdark}{rgb}{0.45, 0.78, 0.75}
\definecolor{teallight}{rgb}{0.82, 0.93, 0.92}
\definecolor{red}{rgb}{1, 0, 0}
\definecolor{wincolor}{rgb}{0.85, 0.0, 0.0}
\definecolor{darkyellow}{rgb}{0.8, 0.8, 0.5}
\definecolor{darkred}{rgb}{0.7, 0.3, 0.3}
\definecolor{darkgreen}{rgb}{0.3, 0.7, 0.3}
\definecolor{green}{rgb}{0, 1.0, 0}
\definecolor{pink}{rgb}{1, 0.4, 0.7}
\definecolor{MyCustomBlue}{HTML}{150CA6}
\usepackage[
    pagebackref,
    breaklinks,
    colorlinks=true,
    citecolor=MyCustomBlue,
    pdftitle={StructParts},
]{hyperref}
\usepackage{xspace}
\newcommand{\StructParts}{{\textsc{StructParts}}\xspace}
\newcommand{\Ours}{\StructParts}

\newcommand{\Point}{\mathbf{X}}

\usepackage{amsmath,amsfonts,bm}

\def\Tabref#1{Tab.~\ref{#1}}
\def\Secref#1{Sec.~\ref{#1}}
\def\Appref#1{App.~\ref{#1}}
\def\Figref#1{Fig.~\ref{#1}}

\def\eqref#1{equation~\ref{#1}}

\def\1{\bm{1}}

\def\rvq{{\mathbf{q}}}

\def\rvz{{\mathbf{z}}}

\def\rmC{{\mathbf{C}}}

\def\rmI{{\mathbf{I}}}

\def\rmV{{\mathbf{V}}}

\def\rmZ{{\mathbf{Z}}}

\DeclareMathAlphabet{\mathsfit}{\encodingdefault}{\sfdefault}{m}{sl}
\SetMathAlphabet{\mathsfit}{bold}{\encodingdefault}{\sfdefault}{bx}{n}

\def\gE{{\mathcal{E}}}
\def\gF{{\mathcal{F}}}

\def\gL{{\mathcal{L}}}

\def\gS{{\mathcal{S}}}

\def\gV{{\mathcal{V}}}

\def\sR{{\mathbb{R}}}

\title{
Seeing Both the Forest and the Trees: Reusing Holistic 3D Priors for Part-Decomposed Generation\thanks{Project page: \url{https://zhao-yiqun.github.io/structparts/}.}
}

\author{%
  Yiqun Zhao$^{1,2}$ \quad Binbin Huang$^{1,2}$ \quad Haobin Duan$^{1,2}$ \quad Zibo Zhao$^{3}$ \quad Shenghua Gao$^{1,2,4}$\thanks{Corresponding author.} \\
  $^1$The University of Hong Kong \quad $^2$Transengram \\
  $^3$Tencent \\
  $^4$Shenzhen Loop Area Institute
}

\begin{document}

\maketitle

\begin{figure*}[h]
    \centering
    \includegraphics[trim=0cm 6cm 0cm 10cm, clip, width=\textwidth]{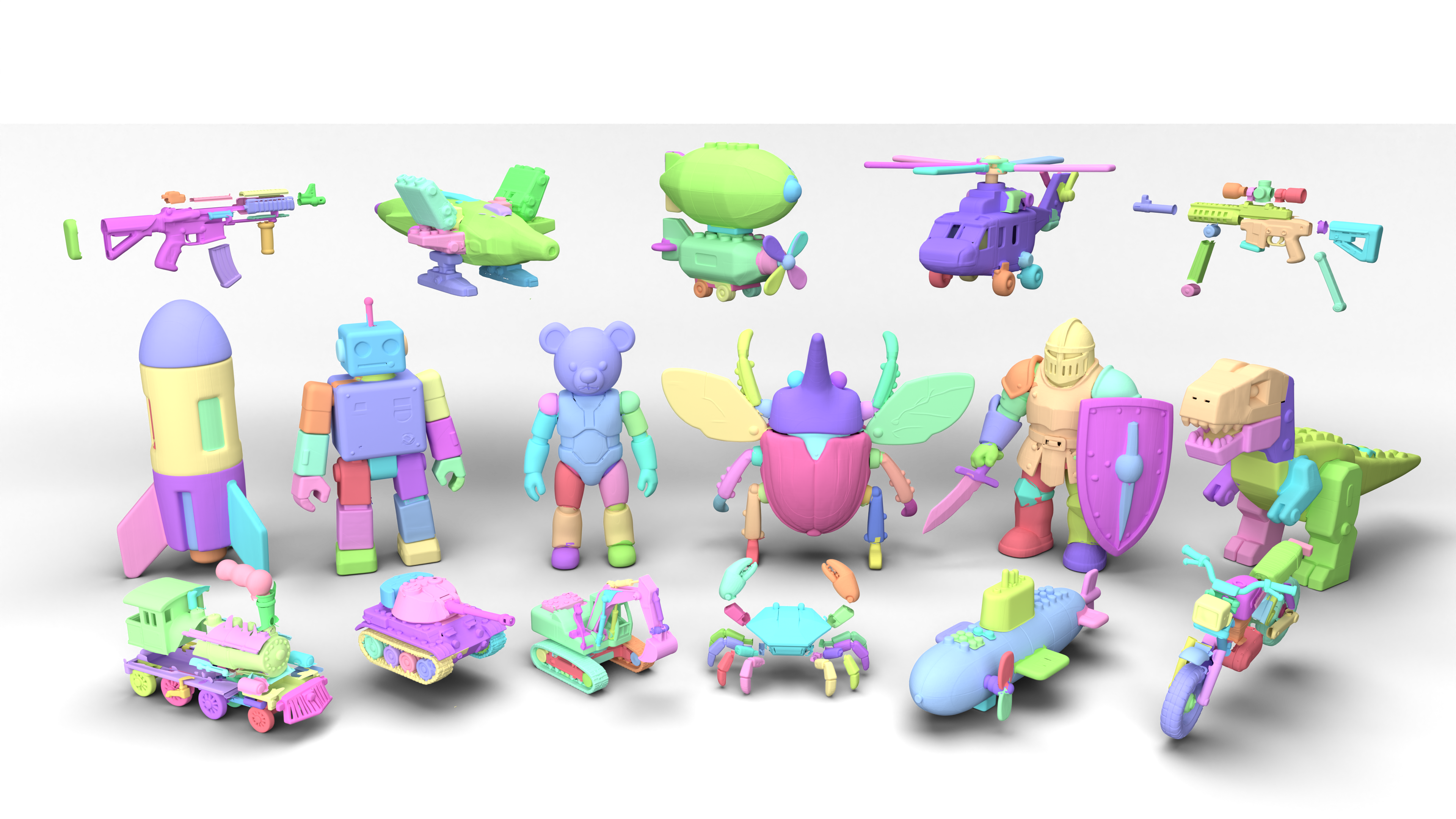}
    \caption{Diverse decomposed 3D objects generated by our approach using images from recent open-source image generation models~\cite{zimage2025, flux2_2025}. Among them, the guns, train, crab, and motorcycle are specifically displayed in exploded views
    }
    \label{fig:teaser}
\end{figure*}


\begin{abstract}

Generating 3D assets with explicit parts requires two properties at once: geometric detail inside each part and structural coherence across the whole object. Recent image-conditioned 3D generation models adopt a \emph{structured 3D representation} that anchors tokens to explicit spatial positions, capturing both properties, but their output remains a single fused mesh. We ask how a part-decomposed generation model can inherit both properties of these holistic priors, and answer with \Ours, a framework whose core technical contribution is a \emph{dual-frame coordinate interleaving} mechanism: each shape latent is interpreted under both a part-local and an object-global reference frame, and processed through a global-local interleaved transformer that simultaneously reuses the prior's geometric detail at the part scale and its structural coherence at the whole-object scale. As its structural input, \Ours uses a \emph{segmented mesh scaffold}, a coarse mesh paired with a face-level segmentation mask kept entirely within the prior's native 3D representation. With only lightweight finetuning, both intra-part detail and inter-part coherence emerge from the same pretrained weights. On PartObjaverse-Tiny, \Ours matches or surpasses prior part-decomposed models at a fraction of their training cost, while naturally supporting part-mesh refinement and part-level editing.

\end{abstract}
\section{Introduction}
\label{sec:intro}

Part-decomposed 3D asset generation is essential for downstream tasks such as localized editing and animation. A part-decomposed generation model must produce two properties at once: \emph{geometric detail} inside each part, and \emph{structural coherence} across the whole object. Recent image-conditioned 3D generation models~\cite{lai2025latticedemocratizehighfidelity3d, xiang2025structured} adopt a \textbf{structured 3D representation} that anchors tokens to explicit spatial positions, jointly scaling to high-fidelity global structure and local geometric detail. The native 3D priors of such holistic models therefore already encode what part-decomposed generation needs, even though they emit a single fused mesh. Yet directly extending these priors to part-decomposed generation has proven non-trivial.

Approaches that adapt holistic 3D generation models to part-decomposed generation can be grouped into three families, and none of them fully inherits both properties of the holistic prior. (i) Methods that feed a point cloud~\cite{yan2025xpart, yang2025holopart, zhang2025bang}, sometimes with extra abstract layout such as bounding boxes~\cite{yan2025xpart}, supply an input the holistic prior was never trained to take; the part model must therefore relearn its conditioning almost from scratch, incurring substantial retraining cost~\cite{yan2025xpart, zhang2025bang}. (ii) Two-stage methods~\cite{yang2025omnipart, fullpart} repurpose the holistic generation model itself to predict structural abstractions such as bounding boxes or cuboid meshes in the first stage; the resulting objective shift drags the prior into a non-geometric latent space and erases the very geometric detail the pretraining was meant to provide. (iii) Methods that keep the original holistic input but predict each part directly in the unnormalized object frame~\cite{lin2025partcrafter, tang2024partpacker} preserve the input format but mismatch the prior's normalized pretraining space at the output side, where tiny parts occupy only a small fraction of the canvas and degrade accordingly. Across all three, the prior is forced to operate outside the conditions it was pretrained on, leaving neither its geometric detail nor its structural coherence fully inherited. This motivates our central question:

\emph{How can a part-decomposed generation model inherit both the structural coherence and the geometric detail of a holistic 3D prior?}

Answering this question requires recognizing a fundamental characteristic of 3D data: its inherent hierarchical structure. A hat is both a component of a human character and a complete shape in its own right. This dual identity is not incidental but ubiquitous, and a holistic 3D prior trained on diverse 3D assets has implicitly observed each part as both a piece and a whole. To honor this duality, a part-decomposed model should not inherit the prior at one scale only: it should expose both scales simultaneously, treating each part as a complete shape locally (so per-part geometric detail is reused exactly as during pretraining) and as part of the whole object globally (so structural coherence is preserved across the assembly). We summarize this principle as seeing both \textbf{the forest and the trees}: just as a tree is itself a complete shape and at the same time part of a larger forest, a model should reason at the part scale and the object scale at once. The duality is mutually reinforcing: global context constrains parts toward coherence, while high-quality parts elevate the geometric fidelity of the whole.

Honoring this duality requires the model to read the prior at both scales simultaneously, which in turn forces a careful choice of structural input. We propose \Ours, whose first design choice is to keep the structural input format that the holistic prior was already pretrained on. Recent holistic 3D generation models~\cite{lai2025latticedemocratizehighfidelity3d, jia2025ultrashape} condition shape synthesis on a coarse 3D mesh via voxel-anchored positional embeddings; we recognize that this same format already carries everything a part model needs to anchor its latents in 3D space, with the addition of a single face-level seg\-men\-ta\-tion mask sufficing to expose part boundaries. Together, the coarse mesh and seg\-men\-ta\-tion mask form a \textbf{segmented mesh scaffold}: a structural input the holistic prior reads without any objective shift, and one that off-the-shelf upstream stages produce as a by-product of holistic generation and segmentation.


With the scaffold in place, the core mechanism of \Ours is \textbf{dual-frame coordinate interleaving}, which inherits both prior properties at once. Each scaffold-anchored latent is interpreted under two complementary frames: a \emph{local} frame normalized within each part, and a \emph{global} frame normalized across the whole object. These yield distinct local and global positional embeddings, processed through a global-local interleaved transformer that alternates between intra-part and inter-part attention. In the local frame, each part is processed as a standalone shape, reusing the prior's geometric detail as during pretraining; in the global frame, the parts are coordinated as one anchored assembly, preserving the prior's structural coherence. Both properties therefore emerge from the same pretrained weights with only lightweight finetuning, rather than requiring extensive retraining~\cite{zhang2025bang, tang2024partpacker, yan2025xpart}.

In practice, our pipeline stays simple. We train on segmented ground-truth watertight meshes; at inference, scaffolds come from a single image via structure generation~\cite{hunyuan3d2025hunyuan3d21} and 3D segmentation models~\cite{ma2025p3sam,partfield2025}. We retain the image condition alongside the scaffold; together with the holistic prior for geometric details, this enables part-level mesh refinement even when the scaffold is coarse. The scaffold also supports part-level editing via merge/split through segmentation regrouping in a training-free manner. On PartObjaverse-Tiny, \Ours matches or surpasses prior part-decomposed models with substantially lower training costs, showing that strong part-decomposed 3D generation can emerge from a minimal reinterpretation of a holistic prior rather than a part-specific model trained from scratch; we demonstrate generation examples from \Ours in \Figref{fig:teaser}.


In summary, our contributions are three-fold:
\begin{itemize}
    \item We propose a \emph{dual-frame coordinate interleaving} mechanism that lets a part-decomposed generation model inherit both the geometric detail and structural coherence of a holistic 3D prior with only lightweight finetuning.
    \item We adapt the coarse-mesh conditioning from holistic 3D generation models by adding a face-level segmentation mask, yielding a segmented mesh scaffold that exposes part boundaries while remaining in the prior's native 3D representation.
    \item On PartObjaverse-Tiny, our method matches or surpasses prior part-aware methods at a fraction of their training cost, while supporting part-mesh refinement and part-level editing.
\end{itemize}

\section{Related Work}
\label{sec:related}

\textbf{3D holistic shape generation.}
Recent 3D shape generation has converged on two representation paradigms. \emph{Sparse-Voxel} methods~\cite{xiang2024structured, xiang2025structured} restrict computation to active surface voxels, yielding a structured latent space well-suited for editing, but the active-voxel sequence grows rapidly with resolution and demands sparse convolution and specialized attention. \emph{VecSet}-based approaches~\cite{zhao2023michelangelo, zhang2024clay, Chen_2025_Dora, yang2024hunyuan3d, hunyuan3d22025tencent, hunyuan3d2025hunyuan3d21} compress a shape into a small set of feature vectors via cross-attention between a dense point cloud and sparser point queries, fitting cleanly into standard transformer attention. Recent localizable VecSet variants such as LATTICE~\cite{lai2025latticedemocratizehighfidelity3d, jia2025ultrashape} anchor each token to an explicit 3D position, inheriting both VecSet's efficiency and Sparse-Voxel's spatial grounding and capturing both global structure and local geometric detail. Despite this progress, all such holistic models still emit a single fused mesh, motivating a part-decomposed counterpart that directly reuses these native 3D priors rather than disrupting them.

\textbf{3D parts reasoning.}
3D part reasoning extracts part structure from existing 3D shapes. \emph{Traditional supervised} methods~\cite{Mo_2019_CVPR} train segmentation networks on labeled benchmarks but struggle to generalize beyond seen categories. \emph{2D-lifting} methods~\cite{xu2025sampro3d, Peng2023OpenScene} render multiple views and propagate masks from 2D foundation models~\cite{kirillov2023segany, oquab2023dinov2} back to the 3D surface; this enables open-vocabulary segmentation but suffers from view inconsistency and unstable post-processing. \emph{Native 3D data-engine} methods~\cite{yang2024sampart3d, partfield2025, zhou2025pointsam, ma2025p3sam} instead distill 2D foundation features into a 3D backbone for direct prediction. None of these synthesize new geometry; they yield part masks on a given mesh. Our work directly exploits this readily available signal at inference time, treating the segmented mesh as a structural scaffold injected into a holistic generation model, rather than training a separate part-prediction head from scratch or coupling decomposition and geometry generation within a single model~\cite{tang2024partpacker, lin2025partcrafter}.

\textbf{3D decomposed shape generation.}
Existing part-decomposed generation models acknowledge native 3D priors but exploit them only indirectly, falling into four patterns. (i) Conditioning a learned model on a point cloud, sometimes with additional abstract layout~\cite{yan2025xpart, yang2025holopart, zhang2025bang}; because such input formats differ from how holistic shape models are conditioned, the prior is not preserved and substantial retraining is required~\cite{yan2025xpart, zhang2025bang}. (ii) Repurposing the generation model to first synthesize structural abstractions like bounding boxes or cuboid meshes~\cite{yang2025omnipart, fullpart}; the first-stage objective shift drags the prior into a latent space sharply different from the native shape prior. (iii) Keeping the original holistic input but predicting each part in the unnormalized object frame~\cite{lin2025partcrafter, tang2024partpacker, chen2025autopartgen}; this spatial-resolution mismatch hurts tiny parts that occupy only a small fraction of the object volume. (iv) Lifting 2D multi-view part predictions to 3D~\cite{chen2024partgen}; this sidesteps the native 3D latent at the cost of view consistency and post-processing complexity. 
In contrast, \Ours stays in the native geometric domain, exposes part boundaries through a segmented mesh scaffold, and uses dual-frame coordinate interleaving to reason at both scales with only lightweight finetuning rather than full retraining of the underlying generation model.
\section{Method}
\label{sec:method}

\begin{figure*}[!t]
    \centering
    \includegraphics[width=\textwidth]{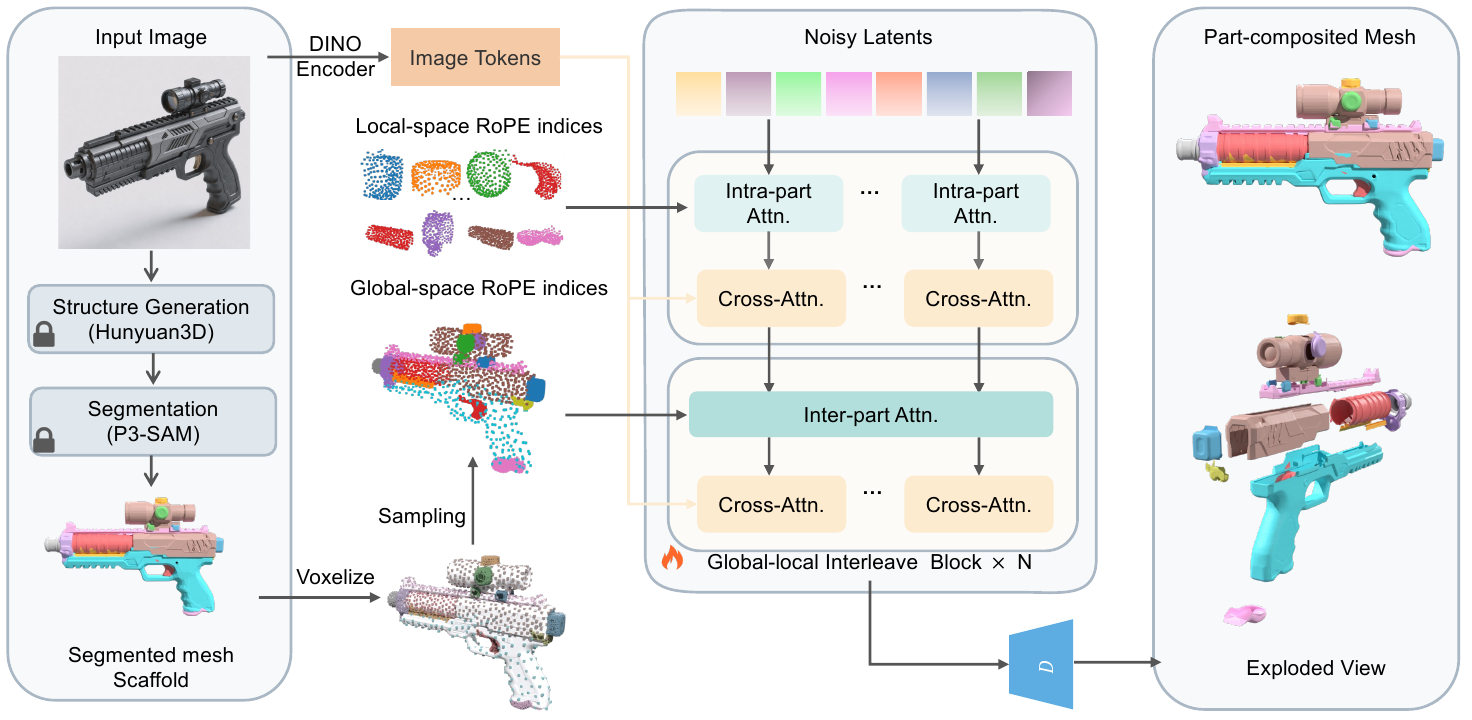}
    \caption{\textbf{Overview of \Ours's pipeline.} Given a single image, our method first constructs a 3D segmented scaffold using the structure generation model followed by a segmentation model. We then sample points from each part surface to construct the voxel indices in both the per-part normalized space and the shared global-shape space. These voxel indices serve as position embeddings to guide the diffusion model's sampling process. The segmented scaffold and part meshes depicted here are dataset-provided ground truth (GT), used to illustrate the part-wise data flow.}
    \label{fig:method}
    \vspace{-1em}
\end{figure*}

This section instantiates \Ours. As illustrated in \Figref{fig:method}, given a single image and a segmented mesh scaffold, we sample per-part shape latents anchored to the scaffold, attach dual-frame positional embeddings (one local to each part, one global across the object), and process them with a global-local interleaved transformer; the image condition is retained throughout, so fine detail flows from the image into each part even when the scaffold is coarse. We review the structured latent representation we build on (\Secref{sec:method_prelim}), detail the scaffold-guided global-local interleaving (\Secref{sec:method_Glocal}), and describe its training-free applications (\Secref{sec:method_applications}).

\subsection{Preliminaries: structured shape latents for 3D diffusion}
\label{sec:method_prelim}

\textbf{Structured shape latents.} We build on recent 3D shape generation models~\cite{lai2025latticedemocratizehighfidelity3d, jia2025ultrashape} that adopt a \emph{structured} 3D representation, in which each latent token is anchored to an explicit spatial position. Given a watertight mesh, the encoder takes a curvature-aware sample~\cite{Chen_2025_Dora, hunyuan3d22025tencent} of point attributes $\Point \in \sR^{N \times 7}$ (positions, normals, sharp-edge flags). We quantize $\Point$ onto a regular grid of resolution $R$, yielding a query set $\Point_{0} \in \sR^{N_{0} \times 7}$ with integer voxel indices $\rmV_{0} \in \{0,\dots,R{-}1\}^{N_{0} \times 3}$ (one token per occupied cell), providing stable spatial information for the 3D DiT's positional embeddings (PE).

The transformer-based VAE encoder maps the shape into these structured latents:
\begin{equation}
 \rmZ = \gE(\Point) = \mathrm{SelfAttn}^{8}\!\bigl(\mathrm{CrossAttn}(\mathrm{PE}(\Point_{0}),\mathrm{PE}(\Point))\bigr),
\end{equation}
where $\mathrm{PE}$ denotes a Fourier positional embedding and $\rmZ = \{\rvz_i\}_{i=1}^{N_0}$ is the set of $N_0$ shape latents. The VAE decoder takes $\rmZ$ and a query point $\rvq$ as input to decode the implicit surface-field value at $\rvq$:
\begin{equation}
    \mathcal{D}(\rvq, \rmZ) = \mathrm{CrossAttn}(\mathrm{PE}(\rvq), \mathrm{SelfAttn}^{16}(\rmZ)).
\end{equation}

\textbf{3D shape diffusion.} Following~\cite{hunyuan3d22025tencent, zhao2023michelangelo}, we apply a flow-based diffusion model~\cite{lipman2024flowmatchingguidecode} on top of these structured latents, with the integer voxel indices $\rmV_0$ providing a Rotary Positional Embedding (RoPE)~\cite{su2021roformer} for each latent token.

\subsection{Scaffold-guided global-local interleaving in transformer}
\label{sec:method_Glocal}

\textbf{Scaffold construction.} Given a single image, we build a \textbf{segmented mesh scaffold} $\gS = (\gV, \gF, \gL)$, where $(\gV, \gF)$ is a coarse mesh produced by an off-the-shelf holistic 3D generation model~\cite{hunyuan3d2025hunyuan3d21} and $\gL$ is a face-level segmentation mask from an off-the-shelf 3D segmentation model such as P3-SAM~\cite{ma2025p3sam}, partitioning the scaffold into $K$ semantic parts. This format extends the coarse-mesh conditioning that holistic 3D models already accept with only a face-level mask, keeping the input in the prior's native 3D representation.

During training, each ground-truth part $k \in \{1, \dots, K\}$ is encoded into a per-part latent $\rmZ^{(k)}$ via the encoder $\gE$ in \Secref{sec:method_prelim}; the full object latent is then the union $\rmZ = \bigcup_{k=1}^K \rmZ^{(k)}$. At inference, per-part latents are initialized from Gaussian noise and generated by the diffusion model, while the segmented scaffold supplies their spatial anchors.

\textbf{Dual-frame positional embeddings.} To read the prior at both scales, we interpret each shape latent under two complementary coordinate frames. For each part $k$, we sample dense points $\Point^{(k)}$ on its scaffold surface, normalize them into a unit cube using the part's bounding box, and voxelize to obtain local voxel indices $\rmI_{L}^{(k)}$; inverting the bounding-box transform lifts the same points to the object's global frame, yielding global voxel indices $\rmI_{G}^{(k)}$. From these we compute two RoPE~\cite{su2021roformer} embeddings:
\begin{equation}
    \text{RoPE}(\mathbf{x}, \rmI) = \mathbf{x} \odot \cos(\mathbf{\Theta}_{\rmI}) + \mathbf{x}^* \odot \sin(\mathbf{\Theta}_{\rmI}),
\end{equation}
where $\rmI$ is a 3D voxel-index input, $\odot$ is element-wise multiplication, $\mathbf{x}^*$ pairs adjacent feature dimensions for rotation, and $\mathbf{\Theta}_{\rmI}$ holds the rotary phases derived from $\rmI$. For the latent tokens in part $k$, we use local embeddings $\text{RoPE}_L(\cdot) \!=\! \text{RoPE}(\cdot, \rmI_{L}^{(k)})$ and global embeddings $\text{RoPE}_G(\cdot) \!=\! \text{RoPE}(\cdot, \rmI_{G}^{(k)})$, which are then injected into the shape latents described next.

\textbf{Global-local interleaving attention.} The dual frames let the network reason about each part as a standalone shape locally while still coordinating parts into one coherent object globally. For every latent token $\rmZ_i \in \rmZ^{(k)}$, we identify its nearest active voxel in the part's normalized frame and assign it both $\text{RoPE}_L$ and $\text{RoPE}_G$. As illustrated in the middle of \Figref{fig:method}, our diffusion transformer alternates layer-by-layer between intra-part and inter-part attention, both implemented through one unified attention operation:
\begin{equation}
    \text{Attn}_F(\rmZ^{(k)}, \rmZ_{kv}) = \mathrm{softmax}\!\Big(\tfrac{\text{RoPE}_F(\sigma_q(\rmZ^{(k)}))\,\text{RoPE}_F(\sigma_k(\rmZ_{kv}))^\top}{\sqrt{d}}\Big)\,\sigma_v(\rmZ_{kv}),
\end{equation}
where $\sigma_q, \sigma_k, \sigma_v$ are the query, key, and value projection layers and $d$ is the per-head hidden dimension. In the \emph{intra-part} attention layers we set $F\!=\!L$ and $\rmZ_{kv}\!=\!\rmZ^{(k)}$, so each part is processed as a standalone shape (\textbf{the trees}), reusing the prior's geometric detail exactly as during pretraining. In the \emph{inter-part} attention layers we set $F\!=\!G$ and $\rmZ_{kv}\!=\!\rmZ$, coordinating the parts as one anchored assembly (\textbf{the forest}) and preserving the prior's structural coherence across the whole object. Both properties therefore emerge from the same pretrained weights, and only the alternating positional injection differs from holistic generation.

\textbf{Image conditioning.} In each block, per-part latents cross-attend (without RoPE) to image features $\rmC_{\text{img}}$ extracted by DINOv2~\cite{oquab2023dinov2}, so fine geometric detail can flow directly from the image into each part even when the scaffold itself is coarse.

\textbf{Training.} We optimize the network with a flow-matching~\cite{lipman2022flow, lipman2024flowmatchingguidecode} objective. Given a ground-truth latent $\rmZ_0$, Gaussian noise $\epsilon \sim \mathcal{N}(\mathbf{0}, \rmI)$, and time step $t \sim \mathcal{U}[0, 1]$, we form the interpolated state $\rmZ_t = t\rmZ_0 + (1-t)\epsilon$ and supervise the network $f_\theta$ to predict the velocity $\rmZ_0 - \epsilon$ from $(\rmZ_t, t, \rmC_{\text{img}})$:
\begin{equation}
    \gL_{\text{FM}} = \mathbb{E}_{t, \rmZ_0, \epsilon} \Big[ \big\| f_\theta(\rmZ_t, t, \rmC_{\text{img}}) - (\rmZ_0 - \epsilon) \big\|_2^2 \Big].
\end{equation}

\subsection{Downstream tasks}
\label{sec:method_applications}
Because \Ours retains the holistic prior intact and is scaffold-guided, two capabilities emerge without any additional training:

\textbf{Part-mesh refinement from a coarse scaffold.} The dual-frame embeddings only require coarse voxel indices, and the image condition is retained at every block. As a result, the model can refine a low-fidelity scaffold into high-quality decomposed parts at inference time, with fine geometric detail recovered directly from the image rather than from the scaffold itself.


\textbf{Part-level editing via scaffold regrouping.} The segmentation mask $\gL$ in the scaffold gives users a direct handle on decomposition granularity. Inspired by RePaint~\cite{Lugmayr_2022_CVPR} and X-Part~\cite{yan2025xpart}, we support training-free \textit{merge} and \textit{split} operations: \textit{merge} fuses adjacent scaffold segments into a single coherent part by relabeling their faces in $\gL$, while \textit{split} subdivides an existing segment into distinct sub-parts. At inference, the modified $\gL$ determines which latents are resampled by the flow-matching process; tokens in unedited regions are kept so that the rest of the object remains consistent.

\section{Experiments}
\label{sec:experiments}


We first detail our implementation (\Secref{sec:implement}) and experimental setup (\Secref{sec:exp_setup}). Next, we evaluate our approach against state-of-the-art methods through comprehensive quantitative and qualitative comparisons (\Secref{sec:exp_main}). We then validate our core design choices via ablation studies (\Secref{sec:exp_ablation}), and finally showcase the versatility of our pipeline through various downstream applications (\Secref{sec:application}).


\subsection{Implementation Details}
\label{sec:implement}

\textbf{Backbone and initialization.} \Ours uses a 21-layer DiT backbone following Hunyuan3D-2.1~\cite{hunyuan3d2025hunyuan3d21}, with self-attention layers alternating between intra-part (local frame, $128^3$ part-normalized RoPE) and inter-part (global frame, $512^3$ shape-global RoPE) positional conditioning. We initialize from open-source Hunyuan3D-2.1~\cite{hunyuan3d2025hunyuan3d21} weights and finetune on part-level data.

\textbf{Training data.} We train on Partverse-XL~\cite{fullpart} ($\sim$36K objects, $\sim$240K parts). Raw part meshes are remeshed to watertight via cuBVH~\cite{tang2024partpacker} and 24 conditioning views per object are rendered following Trellis~\cite{xiang2024structured}; segmented watertight scaffolds for the full object are obtained by baking multi-view rendered segmentation labels onto a watertight remesh of the whole shape (full details in \Appref{sec:suppl_impl}).

\textbf{Training compute.} The DiT is trained for 40K steps on 8 H200 GPUs over $\sim$4 days, for a total allocation of 768 H200-hours. Full network configuration (hidden dim, MoE setup, NTK-aware RoPE base) and hyperparameters (optimizer, precision, gradient clipping, per-part token budget) are deferred to \Appref{sec:suppl_impl}.

\subsection{Experimental Setup}
\label{sec:exp_setup}

We evaluate on PartObjaverse-Tiny~\cite{yang2024sampart3d} (200 high-quality objects with detailed segmentation labels), whose UIDs are strictly excluded from our training set. 

\textbf{Metrics.} We report Chamfer Distance (CD) and F-Score at thresholds 0.05 and 0.10 (F-Score@0.05, F-Score@0.10) at two granularities: \emph{part-level}, where each part is normalized into a $[-1, 1]$ bounding box, and \emph{shape-level}, where the assembled object is normalized into a $[-1, 1]$ bounding box.

\textbf{Settings and baselines.} Because baselines differ in whether they accept layout controls, we use two complementary protocols. \textbf{(i) GT-layout-conditioned, part-level.} Methods that consume a layout (HoloPart~\cite{yang2025holopart}, X-Part~\cite{yan2025xpart}, FullPart~\cite{fullpart}, OmniPart~\cite{yang2025omnipart}) receive ground-truth bounding boxes or segmentation masks, yielding a one-to-one correspondence between predicted and ground-truth parts and a direct part-level geometric assessment. \textbf{(ii) Image-to-decomposed-parts, shape-level.} Direct-generation baselines~\cite{yang2025omnipart, lin2025partcrafter, tang2024partpacker} synthesize parts from a single image with uncontrollable output layouts, so all methods are evaluated on the assembled shape; the layout-dependent methods are run with a standardized predicted pipeline using Hunyuan3D-2.1~\cite{hunyuan3d2025hunyuan3d21} for the scaffold mesh and P3-SAM~\cite{ma2025p3sam} for the part labels.

\textbf{Matched-budget comparison.} In addition to released baseline checkpoints, we retrain X-Part with the same Hunyuan3D-2.1 initialization, Partverse-XL data, 512 tokens per part, and 768-H200-hour allocation as \Ours. \Tabref{tab:main_decomp} reports this result separately; implementation details, training schedules, checkpoint comparisons, and per-object compute are given in \Appref{sec:suppl_matched}.

\subsection{Main Results}
\label{sec:exp_main}

\begin{figure*}[!t]
    \centering
    \includegraphics[width=\textwidth]{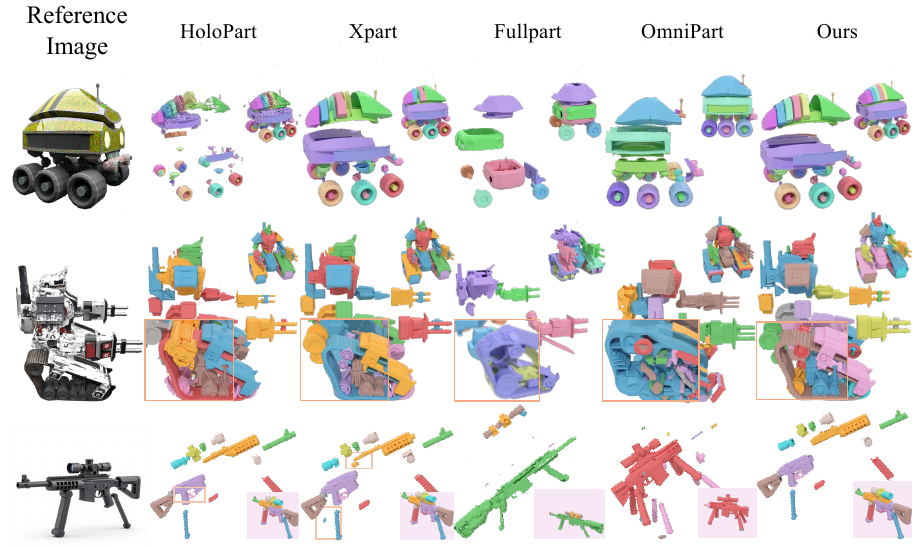}
    \caption{\textbf{Layout-conditioned part decomposition.} Comparison among methods that accept an explicit layout (HoloPart, X-Part, FullPart, OmniPart): each row shows assembled shapes for part coherency and exploded views for per-part geometry. Baselines drop fine details, leave incomplete shapes, or introduce floaters; \Ours preserves both by reusing the holistic prior.}
    \label{fig:exp_1}
\end{figure*}

\textbf{Layout-conditioned evaluation (part-level).} \Tabref{tab:main_decomp} shows \Ours achieves the best decomposition quality across all metrics, with qualitative comparisons in \Figref{fig:exp_1}. The advantage also holds over X-Part retrained under the same GPU-hour budget, providing a controlled comparison of training efficiency. FullPart and OmniPart often fail on complex structures even with ground-truth layouts. HoloPart and XPart follow the given layouts but introduce input signals that disrupt the native 3D prior, yielding missing geometric detail (XPart, row 2), incomplete shapes (row 1 for both), or floaters within part bounding boxes (XPart, row 3). \Ours avoids these failure modes by reusing the holistic prior in its native representation.

\begin{figure*}[!ht]
    \centering
    \includegraphics[width=\textwidth]{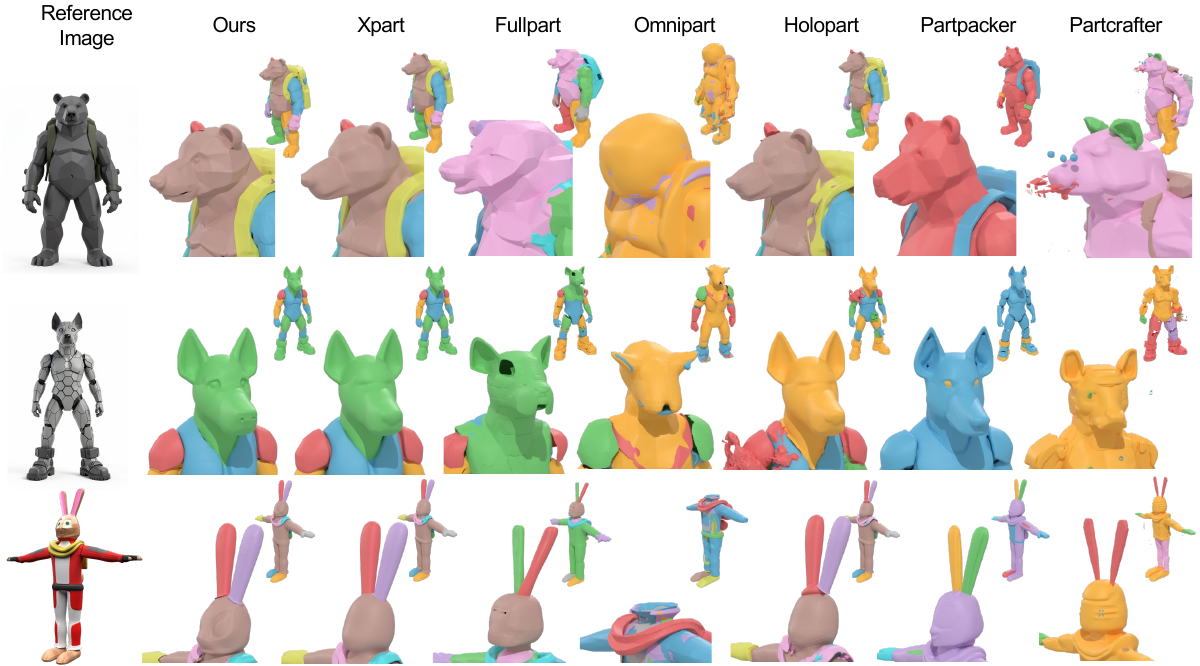}
    \caption{\textbf{Image-to-decomposed-parts comparison.} Top two rows: in-the-wild prompts generated by flux-2~\cite{flux2_2025}; bottom row: an in-distribution test sample. For each input we show the assembled shape and a zoom-in view. PartPacker / PartCrafter struggle to produce semantically meaningful decompositions; FullPart and OmniPart suffer from boundary collisions;}
    \label{fig:image_to_parts}
    \vspace{-1.0em}
\end{figure*}

\textbf{Image-to-decomposed-parts evaluation (shape-level).} \Tabref{tab:main_result} shows \Ours is comparable to released X-Part and outperforms the other baselines by a clear margin; the matched-budget comparison is reported separately in \Tabref{tab:main_decomp}. \Figref{fig:image_to_parts} reveals each baseline's failure mode: PartPacker and PartCrafter cannot produce semantically meaningful decompositions as they synthesize parts without explicit controllability; FullPart and OmniPart's bbox-conditioned formulation causes collisions and mismatches at part boundaries that propagate into the assembled shape. \Ours, XPart and HoloPart share the same scaffold input, yet only \Ours leaves the holistic prior undisturbed, preserving fine geometric detail in the final decomposed shape generation.


\textbf{Additional validation.} On a fixed 200-object ABO subset evaluated with identical inputs and metrics, \Ours obtains the lowest mean and median CD and matches released X-Part at F@0.05. Category-wise results, sample counts, and paired bootstrap intervals are reported in \Appref{sec:suppl_generalization}. A controlled comparison of Hunyuan3D-2.1 and UltraShape initialization at the same training point is given in \Appref{sec:suppl_initialization}.

\begin{table*}[t]
  \centering
  \vspace{0.8em}
  \begin{minipage}[t]{0.48\linewidth}
    \centering
    \caption{\textbf{Part decomposition.} Part-level comparison on PartObjaverse-Tiny. $^{\dagger}$Released X-Part~\cite{yan2025xpart} uses $\sim$2M closed-source objects/128 GPUs versus our $\sim$36K open-source objects/8 GPUs; matched X-Part uses our initialization, data, token and compute budgets (\Appref{sec:suppl_matched}).}
    \label{tab:main_decomp}
    \renewcommand{\arraystretch}{0.9}%
    \resizebox{\linewidth}{!}{%
    \begin{tabular}{lccc}
      \toprule
          Method &  CD $\downarrow$ & F-0.05 $\uparrow$ & F-0.10 $\uparrow$ \\
      \midrule
      OmniPart                 & 0.238 & 0.43 & 0.60 \\
      HoloPart                 & 0.182 & 0.48 & 0.64 \\
      X-Part$^{\dagger}$        & \sbest 0.114 & \sbest 0.63 & \sbest 0.75 \\
      X-Part (matched)         & 0.169 & 0.55 & 0.69 \\
      Fullpart                 & 0.375 & 0.24 & 0.39 \\
      \Ours                    & \best 0.107 & \best 0.67 & \best 0.77 \\
      \bottomrule
    \end{tabular}%
    }
  \end{minipage}
  \hfill 
  \begin{minipage}[t]{0.48\linewidth}
    \centering
    \caption{\textbf{Decomposed shape generation.} Quantitative shape-level comparison on PartObjaverse-Tiny using released baseline checkpoints. $^{\dagger}$The released X-Part model uses the larger training resources noted in \Tabref{tab:main_decomp}; the matched-budget comparison is reported separately in Table~\ref{tab:main_decomp}.}
    \label{tab:main_result}
    \renewcommand{\arraystretch}{0.7}%
    \resizebox{\linewidth}{!}{%
    \begin{tabular}{lccc}
      \toprule
      Method & CD $\downarrow$ & F-0.05 $\uparrow$ & F-0.10 $\uparrow$ \\
      \midrule
      OmniPart                 & 0.088 & 0.56 & 0.75 \\
      PartCrafter              & 0.105 & 0.47 & 0.67 \\
      PartPacker               & 0.068 & 0.56 & 0.76 \\
      HoloPart                 & 0.092 & 0.55 & 0.73 \\
      X-Part$^{\dagger}$        & \best 0.043 & \best 0.65 & \best 0.83 \\
      Fullpart                 & 0.051 & 0.58 & 0.78 \\
      \Ours                    & \sbest 0.045 & \sbest 0.63 & \sbest 0.83 \\
      \bottomrule
    \end{tabular}%
    }
  \end{minipage}
\vspace{-0.8em}
\end{table*}

\subsection{Ablation Studies}
\label{sec:exp_ablation}

\begin{table}[t!]
  \centering
  \caption{\textbf{Ablation Study of dual-frame positional embedding strategies.} Quantitative results evaluated at both part-level and shape-level.}
  \label{tab:ablation}
  \begin{tabular}{lcccccc}
    \toprule
    \multirow{2}{*}{Method} & \multicolumn{3}{c}{Part-level} & \multicolumn{3}{c}{Shape-level} \\
    \cmidrule(lr){2-4} \cmidrule(lr){5-7}
    & CD $\downarrow$ & F-0.05 $\uparrow$ & F-0.10 $\uparrow$ & CD $\downarrow$ & F-0.05 $\uparrow$ & F-0.10 $\uparrow$ \\
    \midrule
    \Ours                 & \best 0.107 & \best 0.67 & \best 0.77 & \best 0.008 & \best 0.93 & \best 0.97 \\
    GRoPE + part embed. & 0.125 & 0.63 & 0.74 & 0.012 & 0.83 & 0.91 \\
    GRoPE only      & 0.143 & 0.63 & 0.75 & 0.019 & 0.70 & 0.90 \\
    LRoPE only       & 0.331 & 0.20 & 0.36 & 0.054 & 0.51 & 0.73 \\
    \bottomrule
  \end{tabular}
  \vspace{-1.5em}
\end{table}
\begin{figure*}[!t]
    \centering
    \includegraphics[width=\linewidth]{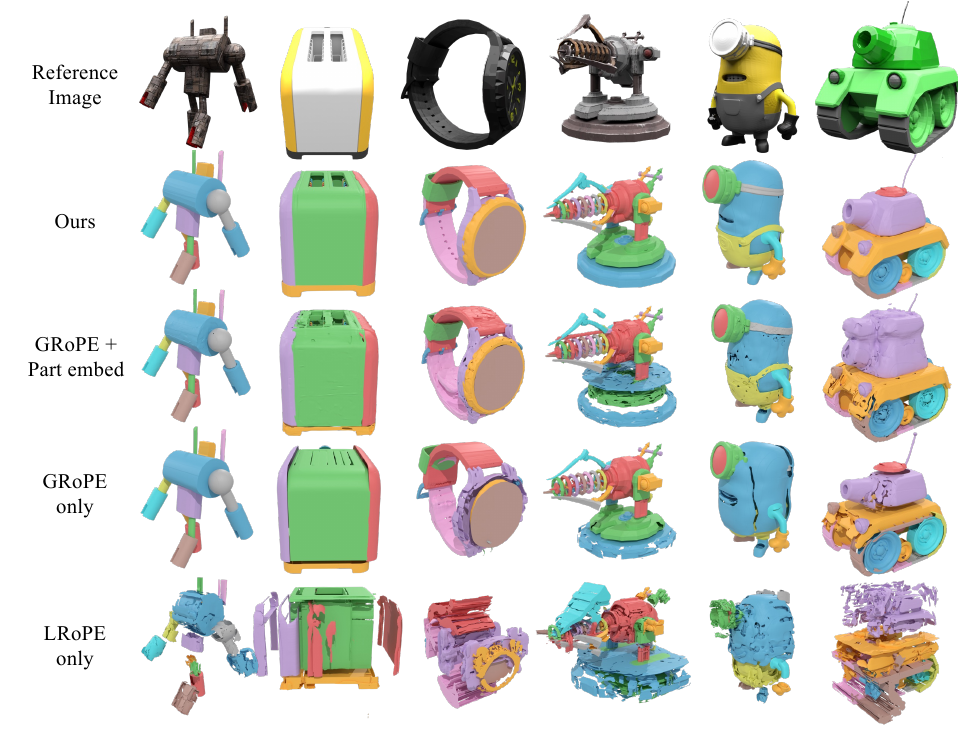}
    \vspace{-2em}
    \caption{\textbf{Ablation on positional-embedding strategies.} Three variants compared against \Ours' dual frame: \emph{LRoPE only} loses global structure (parts treated identically in their own normalized frames); \emph{GRoPE only} recovers global layout but degrades small or off-center parts under unnormalized coordinates; \emph{GRoPE + part embed.} (as in prior work) still misses fine geometric detail. Only the dual-frame design recovers both global structure and per-part detail.}
    \label{fig:ablation}
    \vspace{-0.5em}
\end{figure*}

We validate the dual-frame design in \Tabref{tab:ablation} by asking: (1) is the dual frame itself necessary, and (2) is local RoPE the right local-frame mechanism, compared with the per-part descriptors used in prior work~\cite{yan2025xpart, lin2025partcrafter}? We compare three variants: \textbf{LRoPE only} (local RoPE within each part), \textbf{GRoPE only} (global RoPE across the whole object), and \textbf{GRoPE + part embed.} (global RoPE with per-part embeddings, as in~\cite{yan2025xpart}). \Figref{fig:ablation} answers both. \emph{LRoPE only} treats every part identically in its own normalized frame, severing it from the holistic prior and producing incoherent global structure. \emph{GRoPE only} recovers global layout on simple compositions (top row) but degrades geometry on small or off-center parts under unnormalized coordinates; together these confirm that the dual frame is necessary. Adding per-part embeddings on top of GRoPE still fails to restore fine geometric detail, showing that local RoPE -- not a part descriptor -- is the key driver of per-part geometric quality.




\textbf{Scaffold robustness.} With all other settings fixed, we add normal-direction vertex noise at 1\%, 2\%, and 5\% of the scaffold bounding-box diagonal, or relabel 5\%, 10\%, and 20\% of boundary voxels. Whole-shape performance remains close to clean at 1\% geometry noise, but F@0.05 falls from 0.63 to 0.51 at 5\%. Because assembled-shape metrics can hide incorrect part assignments, we additionally corrupt GT labels and evaluate corresponding parts: F@0.05 remains 0.67 at 5\% corruption but falls to 0.58 at 20\%. Comparing input scaffolds and generated shapes under identical normalization and surface sampling shows comparable overall geometry, without establishing high-frequency detail gains from aggregate metrics alone. Full geometry-preservation and perturbation results are in \Appref{sec:suppl_scaffold}. These tests measure local error sensitivity; substantial natural scaffold failures remain a limitation.

\subsection{Applications}
\label{sec:application}

As discussed in \Secref{sec:method_applications}, the scaffold-conditioned formulation naturally supports downstream tasks. We demonstrate two: decomposed-mesh refinement and part-level editing.


\begin{figure}[!t]
    \centering
    \newlength{\figheight}\setlength{\figheight}{4.2cm}
    \begin{minipage}[t]{0.55\textwidth}
        \centering
        \includegraphics[width=\linewidth,height=\figheight,keepaspectratio]{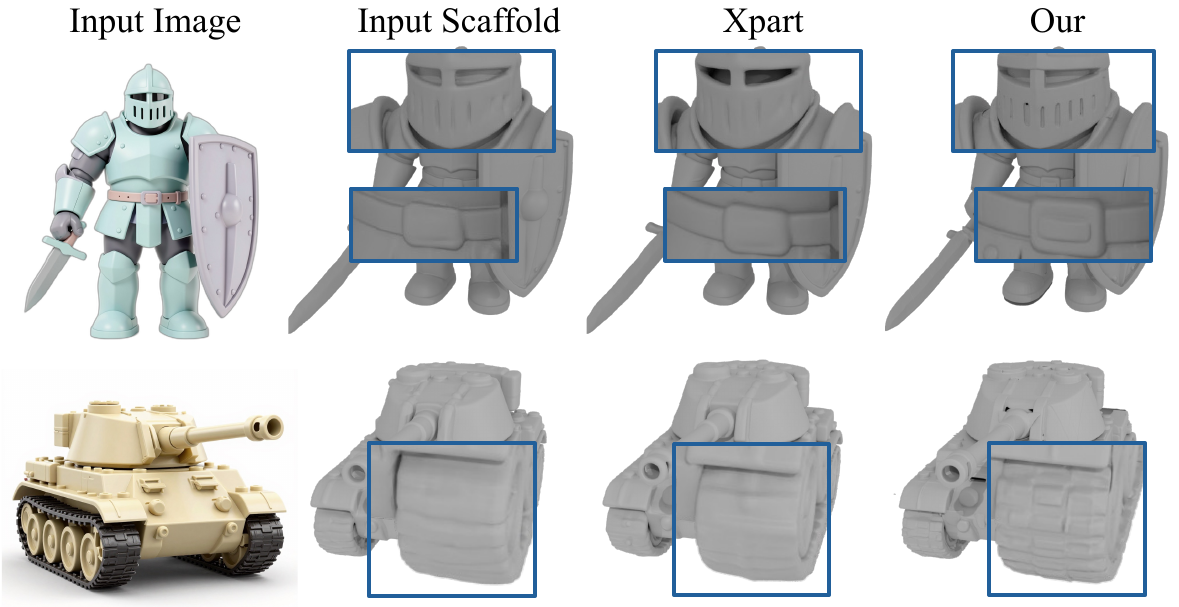}
        \caption{\textbf{Part-mesh refinement.} Given a coarse input scaffold, X-Part rigidly follows the input geometry, while \Ours recovers fine geometric detail directly from the image by leveraging the holistic prior.}
        \label{fig:mesh_refinement}
    \end{minipage}
    \hfill
    \begin{minipage}[t]{0.40\textwidth}
        \centering
        \includegraphics[width=\linewidth,height=\figheight,keepaspectratio]{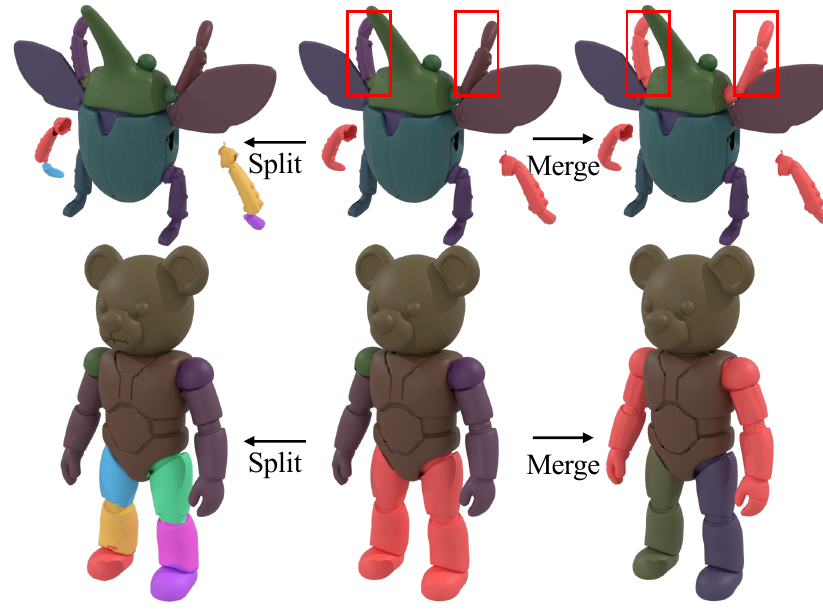}
        \caption{\textbf{Part-level editing.} Merging or splitting segmentation groups on the scaffold yields decompositions at different granularities, training-free at inference time.}
        \label{fig:part_editing}
    \end{minipage}
    \vspace{-1em}
\end{figure}


\textbf{Decomposed-mesh refinement.} As shown in \Figref{fig:mesh_refinement}, when the input scaffold~\cite{hunyuan3d2025hunyuan3d21} is coarse, X-Part~\cite{yan2025xpart} rigidly follows the input geometry and fails to recover detail, while \Ours synthesizes fine geometric details that faithfully match the input image. By aligning each part with the holistic shape latent space, our dual-frame design inherits the detail-recovery ability of the underlying generation prior rather than forcing the model into a new part-specific interface, suggesting potential to scale to high-quality mesh generation.

\textbf{Part-level shape editing.} As shown in \Figref{fig:part_editing}, splitting or merging segmentation groups on the scaffold surface naturally yields part decompositions at different granularities.




\section{Conclusion}
\label{sec:conclusion}

We presented \Ours, a part-decomposed 3D generation framework that inherits both the geometric detail and the structural coherence of a pretrained holistic 3D prior. Its core mechanism, \emph{dual-frame coordinate interleaving}, interprets each shape latent under both a part-local and an object-global reference frame, processed through a global-local interleaved transformer that reads the prior at both scales while remaining in the prior's native 3D representation. The structural input is a \emph{segmented mesh scaffold} borrowed from coarse-mesh conditioning in holistic 3D generation, requiring only a face-level segmentation mask as adaptation. With only lightweight finetuning, \Ours matches or surpasses prior part-decomposed methods on PartObjaverse-Tiny at a fraction of their training cost, and naturally supports part-mesh refinement and part-level editing.

\begin{figure*}[!ht]
    \centering
    \setlength{\tabcolsep}{2pt}
    \begin{tabular}{@{}ccc@{\hspace{8pt}}ccc@{}}
      \parbox[c]{0.15\textwidth}{\centering\footnotesize Input image} &
      \parbox[c]{0.15\textwidth}{\centering\footnotesize Scaffold with segmentation} &
      \parbox[c]{0.15\textwidth}{\centering\footnotesize Part generation} &
      \parbox[c]{0.15\textwidth}{\centering\footnotesize Input image} &
      \parbox[c]{0.15\textwidth}{\centering\footnotesize Scaffold with segmentation} &
      \parbox[c]{0.15\textwidth}{\centering\footnotesize Part generation} \\[4pt]
      \includegraphics[width=0.15\textwidth]{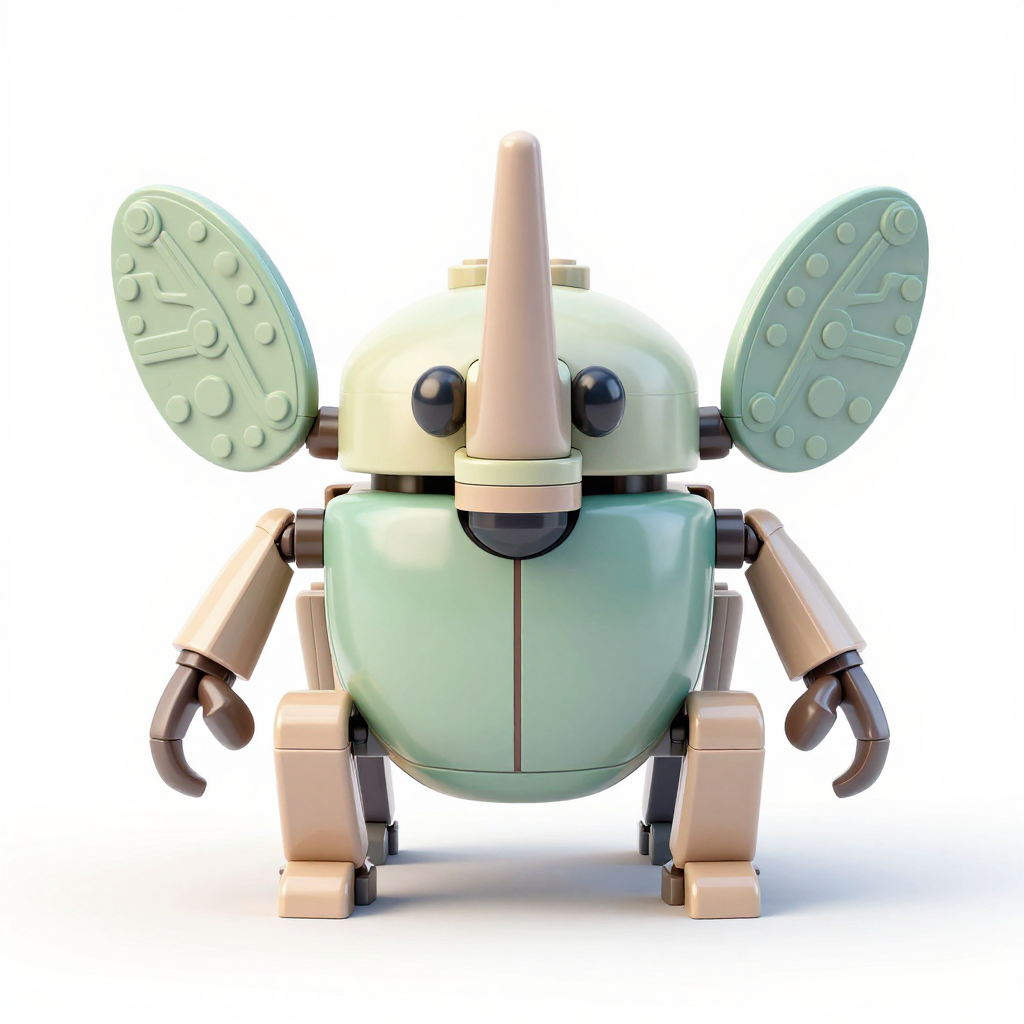} &
      \includegraphics[width=0.15\textwidth]{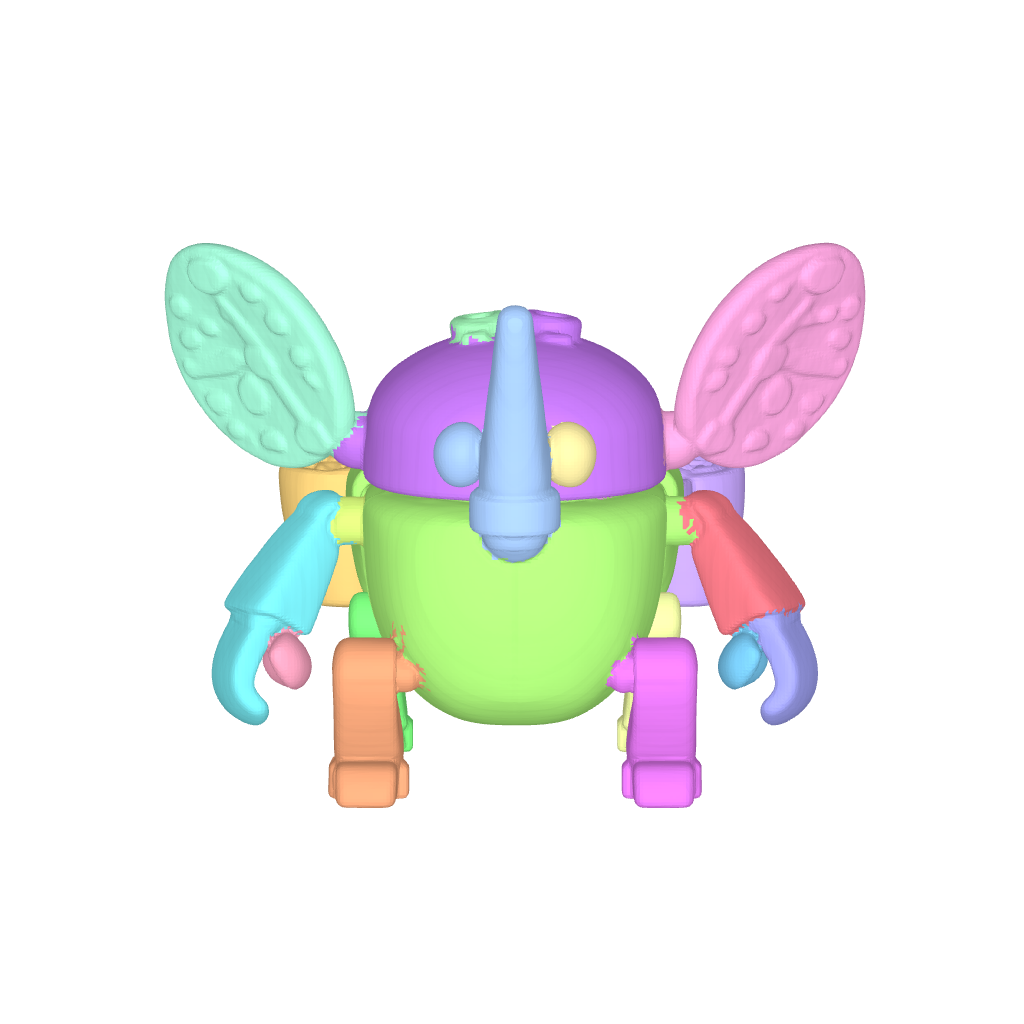} &
      \includegraphics[width=0.15\textwidth]{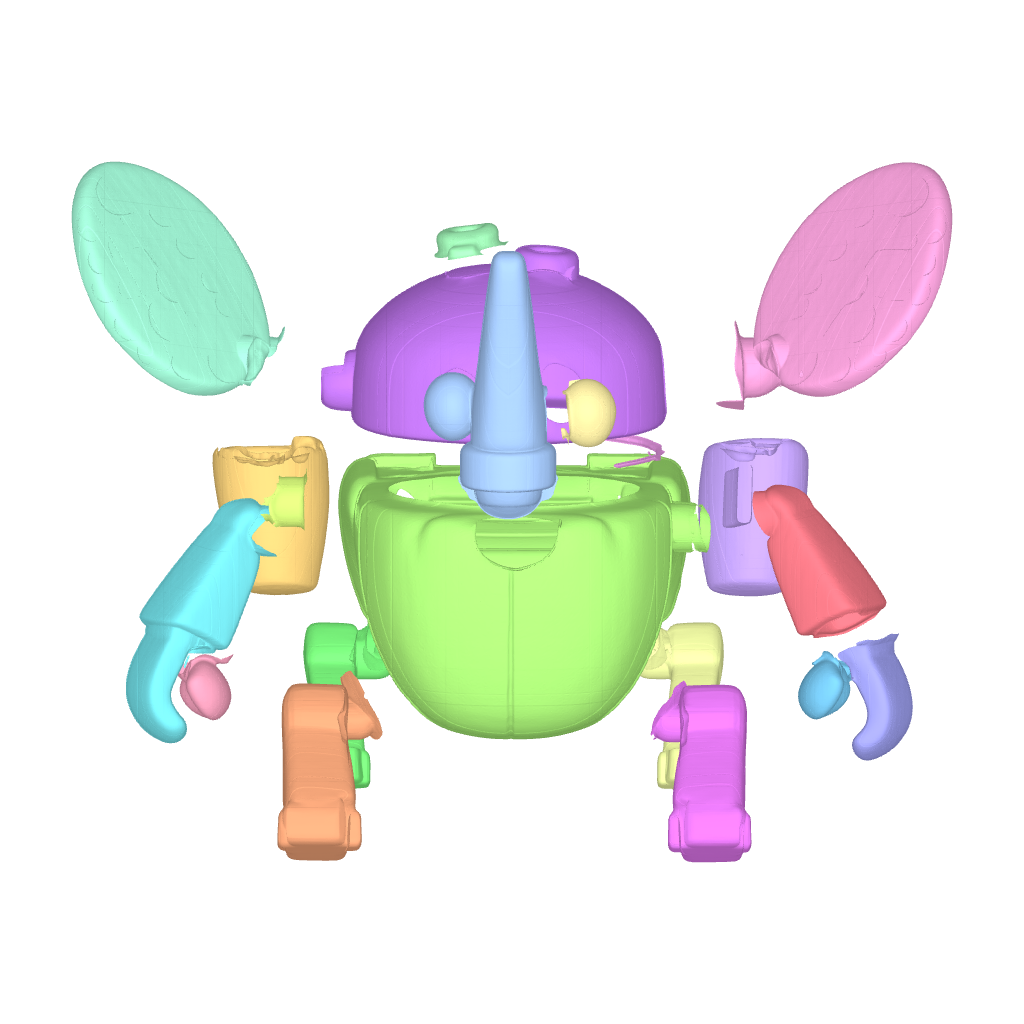} &
      \includegraphics[width=0.15\textwidth]{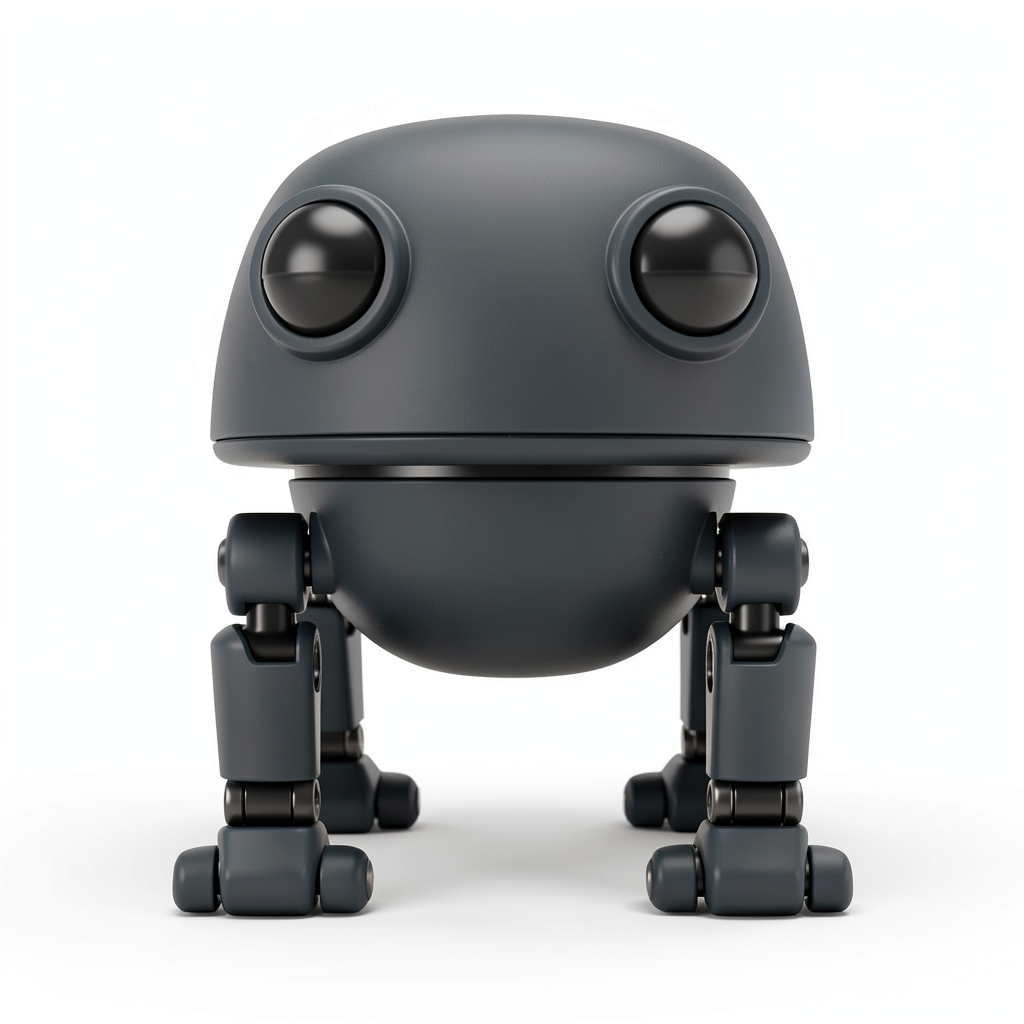} &
      \includegraphics[width=0.15\textwidth]{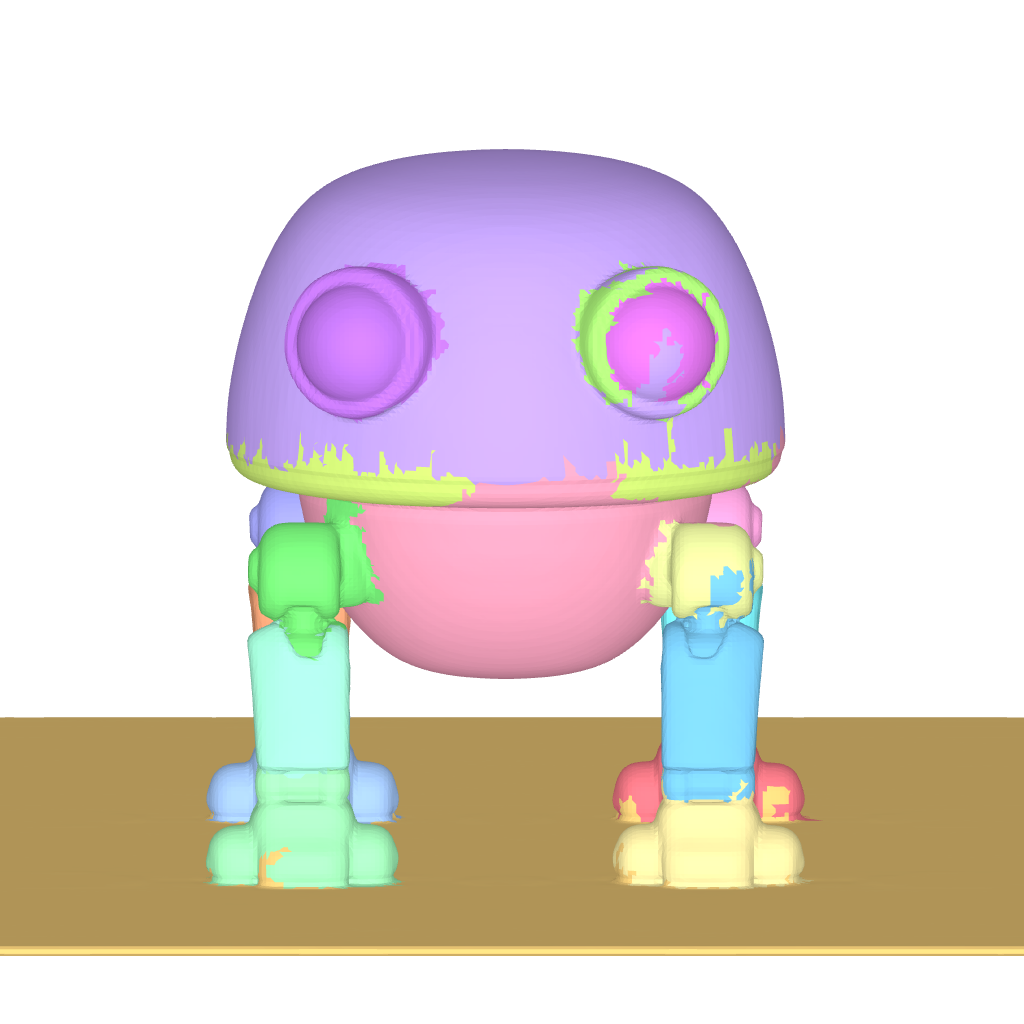} &
      \includegraphics[width=0.15\textwidth]{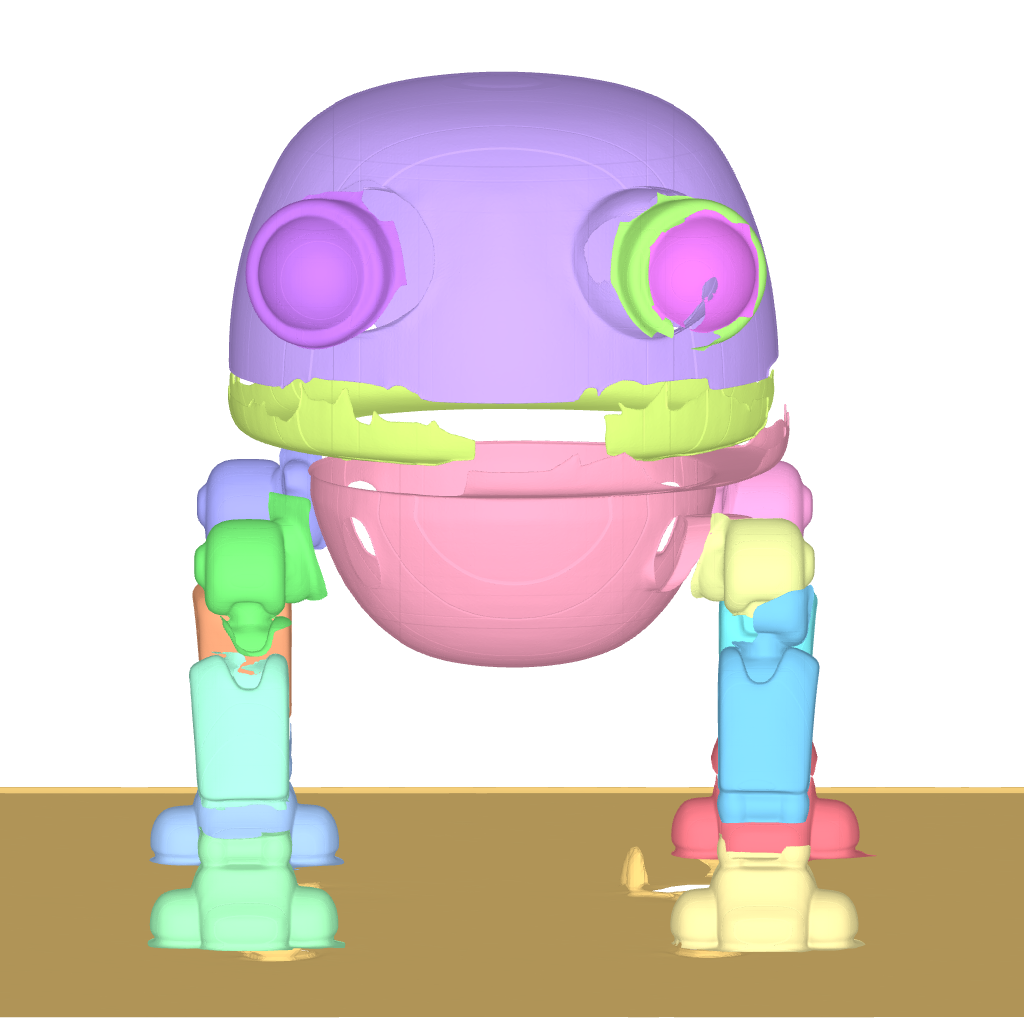}
    \end{tabular}
    \caption{\textbf{Limitations: inaccurate scaffolds.} Each example shows the input image, scaffold with segmentation, and exploded generation. Although our method generates geometric detail, inaccuracies in the input scaffold can persist in the generated parts.}
    \label{fig:limitations}
\end{figure*}

\textbf{Limitations and future work.} \Ours depends on the accuracy of the input scaffold. Although it can recover geometric detail, it does not reliably correct an inaccurate scaffold, and scaffold errors can propagate to the generated parts. The controlled perturbations in \Appref{sec:suppl_scaffold} quantify local sensitivity, rather than the full range of naturally occurring scaffold inaccuracies.
Representative examples of inaccurate scaffolds and their generated parts are shown in \Figref{fig:limitations}.
Promising next steps include extending dual-frame interleaving to deeper hierarchies (e.g., sub-parts) and applying structured-representation reuse to articulated-object generation. Broader societal impacts are discussed in \Appref{sec:suppl_impacts}.

\begin{ack}
This work is partially supported by funding and computing resources provided by the HKU-Shanghai Intelligent Computing Research Center. This work was supported by the General Research Fund of the Research Grants Council of Hong Kong under Grant No. 17200725 and the JC STEM Lab funded by The Hong Kong Jockey Club Charities Trust.
\end{ack}

\bibliographystyle{unsrt}
\bibliography{main}

@String(CVPR= {IEEE Conf. Comput. Vis. Pattern Recog.})

@String(TOG= {ACM Trans. Graph.})

@String(ICLR = {Int. Conf. Learn. Represent.})

@String(CVPR  = {CVPR})

@String(TOG   = {ACM TOG})

@String(ICLR  = {ICLR})

@article{zhang2024clay,
  title={Clay: A controllable large-scale generative model for creating high-quality 3d assets},
  author={Zhang, Longwen and Wang, Ziyu and Zhang, Qixuan and Qiu, Qiwei and Pang, Anqi and Jiang, Haoran and Yang, Wei and Xu, Lan and Yu, Jingyi},
  journal={ACM Transactions on Graphics (TOG)},
  volume={43},
  number={4},
  pages={1--20},
  year={2024},
  publisher={ACM New York, NY, USA}
}

@misc{oquab2023dinov2,
  title={DINOv2: Learning Robust Visual Features without Supervision},
  author={Oquab, Maxime and Darcet, Timothée and Moutakanni, Theo and Vo, Huy V. and Szafraniec, Marc and Khalidov, Vasil and Fernandez, Pierre and Haziza, Daniel and Massa, Francisco and El-Nouby, Alaaeldin and Howes, Russell and Huang, Po-Yao and Xu, Hu and Sharma, Vasu and Li, Shang-Wen and Galuba, Wojciech and Rabbat, Mike and Assran, Mido and Ballas, Nicolas and Synnaeve, Gabriel and Misra, Ishan and Jegou, Herve and Mairal, Julien and Labatut, Patrick and Joulin, Armand and Bojanowski, Piotr},
  journal={arXiv:2304.07193},
  year={2023}
}

@article{lipman2022flow,
  title={Flow matching for generative modeling},
  author={Lipman, Yaron and Chen, Ricky TQ and Ben-Hamu, Heli and Nickel, Maximilian and Le, Matt},
  journal={arXiv preprint arXiv:2210.02747},
  year={2022}
}

@article{zhang2025bang,
  title={BANG: Dividing 3D Assets via Generative Exploded Dynamics},
  author={Zhang, Longwen and Zhang, Qixuan and Jiang, Haoran and Bai, Yinuo and Yang, Wei and Xu, Lan and Yu, Jingyi},
  journal={ACM Transactions on Graphics (TOG)},
  volume={44},
  number={4},
  pages={1--21},
  year={2025},
  publisher={ACM New York, NY, USA}
}

@article{yang2025holopart,
      title={HoloPart: Generative 3D Part Amodal Segmentation},
      author={Yang, Yunhan and Guo, Yuan-Chen and Huang, Yukun and Zou, Zi-Xin and Yu, Zhipeng and Li, Yangguang and Cao, Yan-Pei and Liu, Xihui},
      journal={arXiv preprint arXiv:2504.07943},
      year={2025}
}

@misc{hunyuan3d2025hunyuan3d21,
    title={Hunyuan3D 2.1: From Images to High-Fidelity 3D Assets with Production-Ready PBR Material},
    author={Tencent Hunyuan3D Team},
    year={2025},
    eprint={2506.15442},
    archivePrefix={arXiv},
    primaryClass={cs.CV}
}

@misc{hunyuan3d22025tencent,
    title={Hunyuan3D 2.0: Scaling Diffusion Models for High Resolution Textured 3D Assets Generation},
    author={Tencent Hunyuan3D Team},
    year={2025},
    eprint={2501.12202},
    archivePrefix={arXiv},
    primaryClass={cs.CV}
}

@InProceedings{Chen_2025_Dora,
    author    = {Chen, Rui and Zhang, Jianfeng and Liang, Yixun and Luo, Guan and Li, Weiyu and Liu, Jiarui and Li, Xiu and Long, Xiaoxiao and Feng, Jiashi and Tan, Ping},
    title     = {Dora: Sampling and Benchmarking for 3D Shape Variational Auto-Encoders},
    booktitle = {Proceedings of the Computer Vision and Pattern Recognition Conference (CVPR)},
    month     = {June},
    year      = {2025},
    pages     = {16251-16261}
}

@misc{yang2024hunyuan3d,
    title={Hunyuan3D 1.0: A Unified Framework for Text-to-3D and Image-to-3D Generation},
    author={Tencent Hunyuan3D Team},
    year={2024},
    eprint={2411.02293},
    archivePrefix={arXiv},
    primaryClass={cs.CV}
}

@article{ma2025p3sam,
  title={P3-SAM: Native 3D Part Segmentation},
  author={Ma, Changfeng and Li, Yang and Yan, Xinhao and Xu, Jiachen and Yang, Yunhan and Wang, Chunshi and Zhao, Zibo and Guo, Yanwen and Chen, Zhuo and Guo, Chunchao},
  journal={arXiv preprint arXiv:2509.06784},
  year={2025}
}

@article{yan2025xpart,
  title={X-Part: High Fidelity and Structure Coherent Shape Decomposition},
  author={Yan, Xinhao and Xu, Jiachen and Li, Yang and Ma, Changfeng and Yang, Yunhan and Wang, Chunshi and Zhao, Zibo and Lai, Zeqiang and Zhao, Yunfei and Chen, Zhuo and others},
  journal={CVPR},
  year={2026}
}

@article{yang2025omnipart,
        title={OmniPart: Part-Aware 3D Generation with Semantic Decoupling and Structural Cohesion},
        author={Yang, Yunhan and Zhou, Yufan and Guo, Yuan-Chen and Zou, Zi-Xin and Huang, Yukun and Liu, Ying-Tian and Xu, Hao and Liang, Ding and Cao, Yan-Pei and Liu, Xihui},
        journal={SIGGRAPH Asia},
        year={2025}
}

@article{lin2025partcrafter,
  title={PartCrafter: Structured 3D Mesh Generation via Compositional Latent Diffusion Transformers},
  author={Lin, Yuchen and Lin, Chenguo and Pan, Panwang and Yan, Honglei and Feng, Yiqiang and Mu, Yadong and Fragkiadaki, Katerina},
  year={2025},
  journal={NeuRIPS},
  url={https://arxiv.org/abs/2506.05573}
}

@article{tang2024partpacker,
  title={Efficient Part-Level 3D Object Generation via Dual Volume Packing},
  author={Tang, Jiaxiang and Lu, Ruijie and Li, Zhaoshuo and Hao, Zekun and Li, Xuan and Wei, Fangyin and Song, Shuran and Zeng, Gang and Liu, Ming-Yu and Lin, Tsung-Yi},
  journal={NeuRIPS},
  year={2025}
}

@article{xiang2024structured,
    title   = {Structured 3D Latents for Scalable and Versatile 3D Generation},
    author  = {Xiang, Jianfeng and Lv, Zelong and Xu, Sicheng and Deng, Yu and Wang, Ruicheng and Zhang, Bowen and Chen, Dong and Tong, Xin and Yang, Jiaolong},
    journal = {CVPR},
    year    = {2024}
}

@article{jia2025ultrashape,
    title={UltraShape 1.0: High-Fidelity 3D Shape Generation via Scalable Geometric Refinement},
    author={Jia, Tanghui and Yan, Dongyu and Hao, Dehao and Li, Yang and Zhang, Kaiyi and He, Xianyi and Li, Lanjiong and Chen, Jinnan and Jiang, Lutao and Yin, Qishen and Quan, Long and Chen, Ying-Cong and Yuan, Li},
    journal={arXiv preprint arXiv:2512.21185},
    year={2025}
}

@article{fullpart,
      title={FullPart: Generating each 3D Part at Full Resolution},
      author={Ding, Lihe and Dong, Shaocong and Li, Yaokun and Gao, Chenjian and Chen, Xiao and Han, Rui and Kuang, Yihao and Zhang, Hong and Huang, Bo and Huang, Zhanpeng and Wang, Zibin and Xu, Dan and Xue, Tianfan},
      journal={ICLR},
      year={2026},
}

@inproceedings{partfield2025,
      title={PartField: Learning 3D Feature Fields for Part Segmentation and Beyond}, 
      author={Minghua Liu and Mikaela Angelina Uy and Donglai Xiang and Hao Su and Sanja Fidler and Nicholas Sharp and Jun Gao},
      booktitle={Proceedings of the IEEE/CVF International Conference on Computer Vision},
      year={2025},
}

@article{lai2025latticedemocratizehighfidelity3d,
      title={LATTICE: Democratize High-Fidelity 3D Generation at Scale}, 
      author={Zeqiang Lai and Yunfei Zhao and Zibo Zhao and Haolin Liu and Qingxiang Lin and Jingwei Huang and Chunchao Guo and Xiangyu Yue},
      year={2026},
      journal={CVPR},
      url={https://arxiv.org/abs/2512.03052}, 
}

@misc{lipman2024flowmatchingguidecode,
      title={Flow Matching Guide and Code}, 
      author={Yaron Lipman and Marton Havasi and Peter Holderrieth and Neta Shaul and Matt Le and Brian Karrer and Ricky T. Q. Chen and David Lopez-Paz and Heli Ben-Hamu and Itai Gat},
      year={2024},
      eprint={2412.06264},
      archivePrefix={arXiv},
      primaryClass={cs.LG},
      url={https://arxiv.org/abs/2412.06264}, 
}

@inproceedings{
zhao2023michelangelo,
title={Michelangelo: Conditional 3D Shape Generation based on Shape-Image-Text Aligned Latent Representation},
author={Zibo Zhao and Wen Liu and Xin Chen and Xianfang Zeng and Rui Wang and Pei Cheng and BIN FU and Tao Chen and Gang YU and Shenghua Gao},
booktitle={Thirty-seventh Conference on Neural Information Processing Systems},
year={2023},
url={https://openreview.net/forum?id=xmxgMij3LY}
}

@misc{su2021roformer,
      title={RoFormer: Enhanced Transformer with Rotary Position Embedding}, 
      author={Jianlin Su and Yu Lu and Shengfeng Pan and Bo Wen and Yunfeng Liu},
      year={2021},
      eprint={2104.09864},
      archivePrefix={arXiv},
      primaryClass={cs.CL}
}

@InProceedings{Lugmayr_2022_CVPR,
    author    = {Lugmayr, Andreas and Danelljan, Martin and Romero, Andres and Yu, Fisher and Timofte, Radu and Van Gool, Luc},
    title     = {RePaint: Inpainting Using Denoising Diffusion Probabilistic Models},
    booktitle = {Proceedings of the IEEE/CVF Conference on Computer Vision and Pattern Recognition (CVPR)},
    month     = {June},
    year      = {2022},
    pages     = {11461-11471}
}

@article{yang2024sampart3d,
  title={SAMPart3D: Segment Any Part in 3D Objects},
  author={Yang, Yunhan and Huang, Yukun and Guo, Yuan-Chen and Lu, Liangjun and Wu, Xiaoyang and Lam, Edmund Y and Cao, Yan-Pei and Liu, Xihui},
  journal={arXiv preprint arXiv:2411.07184},
  year={2024}
}

@inproceedings{
      peng2024yarn,
      title={Ya{RN}: Efficient Context Window Extension of Large Language Models},
      author={Bowen Peng and Jeffrey Quesnelle and Honglu Fan and Enrico Shippole},
      booktitle={The Twelfth International Conference on Learning Representations},
      year={2024},
      url={https://openreview.net/forum?id=wHBfxhZu1u}
}

@misc{zimage2025,
      title={Z-Image: An Efficient Image Generation Foundation Model with Single-Stream Diffusion Transformer}, 
      author={{Z-Image Team, Alibaba Group}},
      year={2025},
      eprint={2511.22699},
      archivePrefix={arXiv},
      primaryClass={cs.CV},
      url={https://arxiv.org/abs/2511.22699}
}

@misc{flux2_2025,
      title={FLUX.2: Frontier Visual Intelligence},
      author={{Black Forest Labs}},
      year={2025},
      howpublished={\url{https://bfl.ai/blog/flux-2}},
      note={Accessed: 2026-05-02}
}

@article{
    xiang2025structured,
    title={Native and Compact Structured Latents for 3D Generation},
    author={Xiang, Jianfeng and Chen, Xiaoxue and Xu, Sicheng and Wang, Ruicheng and Lv, Zelong and Deng, Yu and Zhu, Hongyuan and Dong, Yue and Zhao, Hao and Yuan, Nicholas Jing and Yang, Jiaolong},
    journal={Tech report},
    year={2025}
}

@InProceedings{Mo_2019_CVPR,
    author = {Mo, Kaichun and Zhu, Shilin and Chang, Angel X. and Yi, Li and Tripathi, Subarna and Guibas, Leonidas J. and Su, Hao},
    title = {{PartNet}: A Large-Scale Benchmark for Fine-Grained and Hierarchical Part-Level {3D} Object Understanding},
    booktitle = {The IEEE Conference on Computer Vision and Pattern Recognition (CVPR)},
    month = {June},
    year = {2019}
}

@article{kirillov2023segany,
    title={Segment Anything},
    author={Kirillov, Alexander and Mintun, Eric and Ravi, Nikhila and Mao, Hanzi and Rolland, Chloe and Gustafson, Laura and Xiao, Tete and Whitehead, Spencer and Berg, Alexander C. and Lo, Wan-Yen and Doll{\'a}r, Piotr and Girshick, Ross},
    journal={arXiv:2304.02643},
    year={2023}
}

@inproceedings{xu2025sampro3d,
    title={SAMPro3D: Locating SAM Prompts in 3D for Zero-Shot Instance Segmentation},
    author={Mutian Xu and Xingyilang Yin and Lingteng Qiu and Yang Liu and Xin Tong and Xiaoguang Han},
    year={2025},
    booktitle={International Conference on 3D Vision (3DV)}
}

@inproceedings{Peng2023OpenScene,
    title={OpenScene: 3D Scene Understanding with Open Vocabularies},
    author={Peng, Songyou and Genova, Kyle and Jiang, Chiyu Max and Tagliasacchi, Andrea and Pollefeys, Marc and Funkhouser, Thomas},
    booktitle={Proceedings of the IEEE/CVF Conference on Computer Vision and Pattern Recognition (CVPR)},
    year={2023}
}

@inproceedings{zhou2025pointsam,
    title={Point-{SAM}: Promptable 3D Segmentation Model for Point Clouds},
    author={Yuchen Zhou and Jiayuan Gu and Tung Yen Chiang and Fanbo Xiang and Hao Su},
    booktitle={The Thirteenth International Conference on Learning Representations},
    year={2025},
    url={https://openreview.net/forum?id=yXCTDhZDh6}
}

@article{chen2024partgen,
    title={PartGen: Part-level 3D Generation and Reconstruction with Multi-View Diffusion Models},
    author={Minghao Chen and Roman Shapovalov and Iro Laina and Tom Monnier and Jianyuan Wang and David Novotny and Andrea Vedaldi},
    journal={arXiv preprint arXiv:2412.18608},
    year={2024}
}

@article{chen2025autopartgen,
    title={AutoPartGen: Autogressive 3D Part Generation and Discovery},
    author={Minghao Chen and Jianyuan Wang and Roman Shapovalov and Tom Monnier and Hyunyoung Jung and Dilin Wang and Rakesh Ranjan and Iro Laina and Andrea Vedaldi},
    journal={arXiv preprint arXiv:2507.13346},
    year={2025}
}


\appendix

\clearpage
\section{Technical appendices and supplementary material}
\label{sec:suppl}

\subsection{Implementation Details}
\label{sec:suppl_impl}

\textbf{Dataset and scaffold construction.} We train on Partverse-XL~\cite{fullpart}, which contains 36K objects with about 240K parts in total. Because the raw per-part meshes in Partverse-XL are not watertight, we apply cuBVH-based watertight remeshing~\cite{tang2024partpacker} to each part before training. For each object, 24 conditioning views are rendered following Trellis~\cite{xiang2024structured}.

To obtain segmented watertight scaffolds for the full object, we (i) watertight-remesh the entire shape, (ii) bake multi-view rendered segmentation labels from the raw (non-watertight) mesh onto the watertight surface using a multi-view depth-fusion procedure~\cite{xiang2024structured}, and (iii) filter out edge labels that fall outside the ground-truth per-part bounding boxes to clean up boundary noise. The result is a 3D-consistent watertight mesh whose face labels match the original part decomposition.

\textbf{Network architecture.} Our DiT backbone follows Hunyuan3D-2.1~\cite{hunyuan3d2025hunyuan3d21}: 21 layers, hidden dimension 2048, 16 attention heads. Self-attention alternates by layer parity: even layers are intra-part (tokens attend only within their own part) and odd layers are inter-part (tokens attend across all parts). Each token carries a 3D voxel coordinate encoded by two-scale rotary positional embeddings: a part-local $128^3$ grid in intra-part layers and a shape-global $512^3$ grid in inter-part layers. The shape-global RoPE uses NTK-aware scaling~\cite{peng2024yarn, su2021roformer} with rotary base $\theta = 40\text{k}$, a $4\times$ scaling of the default $\theta = 10\text{k}$ chosen to accommodate the larger $512^3$ position range. The last 6 blocks replace the dense feed-forward layer with a Mixture-of-Experts feed-forward (8 experts, top-2 routing) to enhance latent-space capacity. All attention layers use FlashAttention-3 kernels.

\textbf{Training procedure.} The DiT is initialized from open-source Hunyuan3D-2.1's ShapeDiT~\cite{hunyuan3d2025hunyuan3d21} weights and fine-tuned on part-level data using a rectified-flow-matching objective with a linear noise schedule and velocity prediction. We use AdamW with learning rate $1\text{e}{-4}$ and weight decay $10^{-2}$, in bfloat16 mixed precision (forward/backward in bf16; master weights and optimizer states in fp32) with gradient-norm clipping at $1.0$. During training we fix the per-part token budget to 512, though the model can scale to more tokens at inference. The DiT is trained on 8 GPUs for 40K steps within a 768-H200-hour allocation (8 H200 GPUs for four days).

\textbf{VAE finetuning.} The VAE component is finetuned with a token length of 512 to ensure it can encode and decode individual parts effectively at the per-part scale used during DiT training.

\textbf{Inference pipeline.} At inference, the segmented mesh scaffold is obtained from a single image via two off-the-shelf upstream stages: a holistic 3D generation model~\cite{hunyuan3d2025hunyuan3d21} produces the coarse mesh, and a 3D segmentation model~\cite{ma2025p3sam} provides the per-face part labels. The image is encoded by DINOv2~\cite{oquab2023dinov2}; per-part shape latents are sampled by the rectified-flow scheduler conditioned on dual-frame RoPE indices derived from the scaffold and on the image features. Because the dual-frame embeddings only require coarse voxel indices and the image condition is retained in every block, low-fidelity scaffolds are sufficient: missing geometric detail is recovered directly from the image (see the part-mesh refinement application in \Secref{sec:method_applications}).

\subsection{Matched-Budget Comparison with X-Part}
\label{sec:suppl_matched}

To verify training efficiency under a controlled resource budget, we retrain X-Part~\cite{yan2025xpart} using the same initialization, training data, and token budget as \Ours (\Secref{sec:exp_setup}). Both runs use 8 H200 GPUs for four days. The X-Part implementation and hyperparameters follow the setup confirmed with its authors; the retrained model is distinguished from its released checkpoint throughout the evaluation.

\textbf{Computation and conditioning.} To measure the cost of the two adaptation strategies, we profile both methods with the same PyTorch profiler, including forward and backward computation. \Tabref{tab:suppl_compute} summarizes the shared training allocation and per-object costs. \Ours uses 32.5\% fewer TFLOPs per object and less memory. X-Part uses learned global-object and per-part local-geometry conditioners, fuses semantic features into both, and injects two 4,096-token streams through separate cross-attention branches in every DiT block. \Ours instead derives dual-frame coordinates from scaffold voxel indices and reuses object-level image conditioning across parts. Computing X-Part conditioning online per part partly explains its approximately $1.5\times$ longer step time. We retain the author-confirmed online point-sampling setup, since fixing points offline can introduce a train--inference mismatch.

\begin{table}[!htbp]
  \centering
  \caption{\textbf{Training allocation and computational cost.} Both methods use 512 tokens per part and the same GPU-hour allocation. TFLOPs/object include forward and backward computation under the same PyTorch profiler. Optimizer steps and processed-part counts are reported with checkpoint quality in \Tabref{tab:suppl_checkpoints}.}
  \label{tab:suppl_compute}
  \resizebox{0.7\linewidth}{!}{%
  \begin{tabular}{lcc}
    \toprule
    Quantity & \Ours & X-Part (matched) \\
    \midrule
    GPUs & 8 H200 & 8 H200 \\
    Total allocation & 768 GPU-hours & 768 GPU-hours \\
    \midrule
    Trainable parameters & 3.051 B & 3.315 B \\
    TFLOPs/object & 92.7 & 137.4 \\
    Allocated GPU memory & 82.71 GB & 115 GB \\
    \bottomrule
  \end{tabular}%
  }
\end{table}

\textbf{Checkpoint comparison.} To verify whether the quality advantage requires more optimization steps, we evaluate the 20K and 40K \Ours checkpoints alongside the 28K matched-budget X-Part model using the same part-level protocol. \Tabref{tab:suppl_checkpoints} reports quality, optimizer steps, and processed-part counts together. Even the 20K \Ours checkpoint performs better on all four metrics after fewer steps and fewer processed parts; quality improves further at 40K. The full-budget comparison supports training efficiency, while the earlier checkpoint shows that the advantage does not require a higher optimizer-step count.

\begin{table}[!htbp]
  \centering
  \caption{\textbf{Checkpoint quality and training-schedule accounting.} Part-level evaluation on PartObjaverse-Tiny. The 40K \Ours and 28K X-Part runs use the same GPU-hour allocation; the 20K \Ours checkpoint is included to compare quality after fewer optimizer steps and processed parts.}
  \label{tab:suppl_checkpoints}
  \resizebox{0.8\linewidth}{!}{%
  \begin{tabular}{lcccccc}
    \toprule
    Run & Steps & Processed parts & Mean CD $\downarrow$ & Median CD $\downarrow$ & F@0.05 $\uparrow$ & F@0.10 $\uparrow$ \\
    \midrule
    \Ours & 20K & 2.56M & 0.129 & 0.013 & 0.64 & 0.76 \\
    X-Part (matched) & 28K & 3.59M & 0.169 & 0.028 & 0.55 & 0.69 \\
    \Ours & 40K & 5.12M & 0.107 & 0.014 & 0.67 & 0.77 \\
    \bottomrule
  \end{tabular}%
  }
\end{table}

\subsection{Scaffold Robustness and Geometry Preservation}
\label{sec:suppl_scaffold}

\textbf{Geometry preservation and whole-shape sensitivity.} To verify geometry preservation and quantify sensitivity to upstream errors, we evaluate the input Hunyuan scaffold and generated outputs against the same GT objects with identical normalization and surface sampling. \Tabref{tab:suppl_scaffold} reports the input scaffold, clean output, and outputs under two controlled perturbations. With all other settings fixed, we add normal-direction vertex noise at 1\%, 2\%, or 5\% of the scaffold bounding-box diagonal, or relabel 5\%, 10\%, or 20\% of boundary voxels. The clean output preserves overall geometry; aggregate metrics do not isolate high-frequency detail, which is assessed qualitatively in \Figref{fig:mesh_refinement}. Geometry noise causes little change at 1\%, moderate degradation at 2\%, and substantial degradation at 5\%. Label corruption changes whole-shape metrics only slightly, indicating that assembly can conceal incorrect part boundaries.

\begin{table}[!htbp]
  \centering
  \caption{\textbf{Geometry preservation and scaffold perturbations.} The input scaffold and generated outputs are evaluated against the same GT shapes with identical normalization and surface sampling. Noise is normal-direction vertex displacement as a percentage of the bounding-box diagonal; label corruption reassigns boundary voxels. All other inference settings are fixed.}
  \label{tab:suppl_scaffold}
  \resizebox{0.8\linewidth}{!}{%
  \begin{tabular}{lcccc}
    \toprule
    Method / condition & Mean CD $\downarrow$ & Median CD $\downarrow$ & F@0.05 $\uparrow$ & F@0.10 $\uparrow$ \\
    \midrule
    Input Hunyuan scaffold & 0.049 & 0.038 & 0.63 & 0.84 \\
    \Ours (clean) & 0.045 & 0.037 & 0.63 & 0.83 \\
    Geometry noise 1\% & 0.046 & 0.037 & 0.63 & 0.83 \\
    Geometry noise 2\% & 0.047 & 0.039 & 0.61 & 0.80 \\
    Geometry noise 5\% & 0.052 & 0.041 & 0.51 & 0.73 \\
    Label corruption 5\% & 0.045 & 0.037 & 0.62 & 0.83 \\
    Label corruption 10\% & 0.046 & 0.037 & 0.61 & 0.82 \\
    Label corruption 20\% & 0.047 & 0.036 & 0.61 & 0.82 \\
    \bottomrule
  \end{tabular}%
  }
\end{table}

\textbf{Part-level sensitivity.} To expose segmentation errors hidden by assembly, we apply the same label-corruption rates to GT part labels and compare each generated part with its corresponding GT part. GT labels retain a known correspondence, whereas P3-SAM part IDs cannot be reliably matched to GT part IDs. \Tabref{tab:suppl_label_parts} shows nearly unchanged performance at 5\% corruption and progressively worse part consistency at 10--20\%. These controlled perturbations assess local error sensitivity; the dependence on reasonable coarse topology and the effects of natural structural failures are discussed in \Secref{sec:conclusion}.

\begin{table}[!htbp]
  \centering
  \caption{\textbf{Part-level sensitivity to segmentation errors.} We corrupt GT boundary labels to retain a known correspondence between each generated and GT part. These results are distinct from the assembled-shape experiment in \Tabref{tab:suppl_scaffold}.}
  \label{tab:suppl_label_parts}
  \resizebox{0.8\linewidth}{!}{%
  \begin{tabular}{lcccc}
    \toprule
    Method / condition & Mean CD $\downarrow$ & Median CD $\downarrow$ & F@0.05 $\uparrow$ & F@0.10 $\uparrow$ \\
    \midrule
    Clean & 0.107 & 0.014 & 0.67 & 0.77 \\
    Label corruption 5\% & 0.108 & 0.014 & 0.67 & 0.77 \\
    Label corruption 10\% & 0.113 & 0.019 & 0.61 & 0.73 \\
    Label corruption 20\% & 0.117 & 0.038 & 0.58 & 0.71 \\
    \bottomrule
  \end{tabular}%
  }
\end{table}

\subsection{Cross-Dataset and Category-Wise Evaluation}
\label{sec:suppl_generalization}

\textbf{Cross-dataset evaluation.} To verify generalization beyond PartObjaverse-Tiny, we evaluate a fixed 200-object ABO subset using identical inputs and the same whole-shape CD/F-score protocol for all methods. \Tabref{tab:suppl_abo} separates released and matched-budget X-Part checkpoints. \Ours achieves the lowest mean and median CD, matches released X-Part at F@0.05, and is within 0.01 at F@0.10.

\begin{table}[!htbp]
  \centering
  \caption{\textbf{Cross-dataset evaluation on ABO.} A fixed 200-object subset is evaluated with identical inputs and the same whole-shape CD/F-score protocol for all methods. Released and matched-budget X-Part checkpoints are distinguished.}
  \label{tab:suppl_abo}
  \resizebox{0.8\linewidth}{!}{%
  \begin{tabular}{lcccc}
    \toprule
    Method / condition & Mean CD $\downarrow$ & Median CD $\downarrow$ & F@0.05 $\uparrow$ & F@0.10 $\uparrow$ \\
    \midrule
    \Ours & \textbf{0.021} & \textbf{0.012} & \textbf{0.68} & 0.79 \\
    X-Part (released) & 0.023 & 0.013 & \textbf{0.68} & \textbf{0.80} \\
    X-Part (matched) & 0.045 & 0.037 & 0.58 & 0.69 \\
    HoloPart & 0.026 & 0.018 & 0.62 & 0.70 \\
    FullPart & 0.028 & 0.021 & 0.66 & 0.71 \\
    OmniPart & 0.095 & 0.057 & 0.49 & 0.49 \\
    \bottomrule
  \end{tabular}%
  }
\end{table}

\textbf{Category consistency and data separation.} To assess whether the gain is concentrated in a few categories, we report part-level CD for the six largest evaluation categories in \Tabref{tab:suppl_category_cd}, and object counts, paired-part counts, and F-scores in \Tabref{tab:suppl_category_f}. \Ours has the lowest median CD in all six categories, the lowest mean CD in five, and leads both F-scores in four. The higher Animals mean nevertheless indicates a heavy-tailed failure. All evaluation UIDs are excluded from training, yielding zero object overlap; PartVerse-XL lacks an aligned taxonomy, so semantic-category overlap cannot be measured objectively.

\begin{table}[!htbp]
  \centering
  \caption{\textbf{Category-wise part-level CD on PartObjaverse-Tiny.} Each cell reports mean / median CD, using the six largest evaluation categories. X-Part denotes its released checkpoint.}
  \label{tab:suppl_category_cd}
  \resizebox{0.8\linewidth}{!}{%
  \begin{tabular}{lccccc}
    \toprule
    Category & FullPart & HoloPart & OmniPart & X-Part & \Ours \\
    \midrule
    Animals & 0.395/0.317 & 0.120/0.031 & 0.182/0.205 & \textbf{0.040}/0.013 & 0.095/\textbf{0.009} \\
    Buildings \& Outdoor & 0.483/0.389 & 0.166/0.068 & 0.403/0.386 & 0.100/0.027 & \textbf{0.068/0.013} \\
    Daily-used & 0.259/0.122 & 0.199/0.043 & 0.229/0.191 & 0.119/0.024 & \textbf{0.066/0.011} \\
    Electronics & 0.453/0.256 & 0.150/0.061 & 0.165/0.146 & 0.152/0.029 & \textbf{0.137/0.022} \\
    Human-shaped & 0.332/0.273 & 0.081/0.028 & 0.177/0.148 & 0.049/0.011 & \textbf{0.041/0.009} \\
    Transportation & 0.377/0.238 & 0.181/0.065 & 0.332/0.211 & 0.208/0.044 & \textbf{0.144/0.022} \\
    \bottomrule
  \end{tabular}%
  }
\end{table}
\begin{table}[!htbp]
  \centering
  \caption{\textbf{Category counts and paired-part F-scores.} Object counts describe the evaluation set; paired parts have complete matched predictions. F-score pairs are F@0.05 / F@0.10. X-Part denotes its released checkpoint.}
  \label{tab:suppl_category_f}
  \resizebox{0.8\linewidth}{!}{%
  \begin{tabular}{lcccc}
    \toprule
    Category & Objects & Paired parts & \Ours & X-Part \\
    \midrule
    Human-Shape & 29 & 423 & 0.74/0.83 & 0.74/0.85 \\
    Animals & 23 & 344 & \textbf{0.74}/0.82 & 0.72/0.84 \\
    Daily-Use & 25 & 335 & \textbf{0.71/0.81} & 0.62/0.74 \\
    Buildings \& Outdoor & 25 & 370 & \textbf{0.67/0.79} & 0.61/0.72 \\
    Transportation & 38 & 615 & \textbf{0.59/0.70} & 0.52/0.64 \\
    Electronics & 34 & 523 & \textbf{0.63/0.74} & 0.58/0.71 \\
    \bottomrule
  \end{tabular}%
  }
\end{table}

\textbf{Paired uncertainty estimates.} To assess statistical reliability against released X-Part, we perform a paired bootstrap over objects with complete matched predictions. We resample objects with replacement 10,000 times, retaining all paired parts within each sampled object. \Tabref{tab:suppl_bootstrap} reports \Ours-minus-X-Part differences and 95\% intervals; negative CD and positive F-score differences favor \Ours. Only the F@0.05 interval excludes zero. Statistical conclusions use the unrounded intervals; displayed F-score values use two decimals.

\begin{table}[!htbp]
  \centering
  \caption{\textbf{Paired bootstrap differences against released X-Part.} Objects with complete paired predictions are resampled 10,000 times, retaining all paired parts within each sampled object. Differences are \Ours minus X-Part. F-score estimates and interval bounds use two decimals; a negative bound that rounds to zero retains its sign. Statistical conclusions use the unrounded intervals.}
  \label{tab:suppl_bootstrap}
  \resizebox{0.6\linewidth}{!}{%
  \begin{tabular}{lcc}
    \toprule
    Metric difference & Estimate & 95\% paired bootstrap interval \\
    \midrule
    Mean CD & $-0.0089$ & $[-0.0198,\,+0.0009]$ \\
    F@0.05 & $+0.04$ & $[+0.02,\,+0.06]$ \\
    F@0.10 & $+0.03$ & $[-0.00,\,+0.05]$ \\
    \bottomrule
  \end{tabular}%
  }
\end{table}

\subsection{Initialization Dependence and Method Comparison}
\label{sec:suppl_initialization}

\textbf{Alternative initialization.} To verify dependence on the Hunyuan3D-2.1 prior, we also initialize \Ours from UltraShape~\cite{jia2025ultrashape}, whose DiT is pretrained with a coarse-mesh anchor. Since the compatible UltraShape DiT requires retraining, we compare both initializations at the same 20K halfway training point using part-level metrics. UltraShape achieves mean/median CD of 0.123/0.015 and F@0.05/F@0.10 of 0.64/0.77; the Hunyuan3D-2.1 baseline is the 20K \Ours checkpoint in \Tabref{tab:suppl_checkpoints}. The nearly identical F-scores support adaptation of both initializations at this common training point.

\textbf{Comparison with UniPart.} To clarify the distinction from dual-space part representations, we compare the two formulations. UniPart first generates and segments a whole-object geometry--segmentation latent, then predicts separate global and canonical part latents through two-stage diffusion. \Ours instead processes one latent within one DiT under alternating part-local and object-global frames, accepting an editable segmented scaffold for training-free merge/split editing. This comparison is conceptual; UniPart code was unavailable at the time of evaluation, preventing a matched quantitative comparison.

\subsection{Social Impact}
\label{sec:suppl_impacts}

\textbf{Positive impact.} \Ours produces explicitly part-decomposed 3D assets, enabling part-aware downstream workflows such as localized editing, animation rigging, retopology, and semantic asset reuse. By reusing a pretrained holistic 3D prior with only lightweight finetuning, the method avoids the energy and data costs of training a part-aware generative model from scratch, lowering the carbon footprint of part-decomposed 3D research and making this capability accessible to teams without large-scale compute infrastructure. The scaffold-conditioned interface and training-free editing operations (merging and splitting parts via segmentation regrouping) further lower the barrier for non-experts to iterate on 3D content, with potential applications in education, indie game development, and rapid prototyping for design and engineering.

\textbf{Negative impact and mitigation.} Like other generative 3D models, \Ours can be misused, for example to produce unauthorized reproductions of copyrighted 3D assets or to fabricate misleading 3D media. Because the model inherits a pretrained holistic prior, it also inherits any biases present in the upstream training data, such as over-representation of common object categories or culturally skewed design styles. To mitigate these risks, we train and release our model only with publicly available, license-compliant datasets, and we encourage downstream users to attribute reused assets and to verify generated outputs for intellectual-property and ethical concerns before deployment. The model is not designed for, and we do not recommend its use in, identity-related or surveillance applications.


\newpage
\section*{NeurIPS Paper Checklist}

The checklist is designed to encourage best practices for responsible machine learning research, addressing issues of reproducibility, transparency, research ethics, and societal impact. Do not remove the checklist: {\bf The papers not including the checklist will be desk rejected.} The checklist should follow the references and follow the (optional) supplemental material.  The checklist does NOT count towards the page
limit. 

Please read the checklist guidelines carefully for information on how to answer these questions. For each question in the checklist:
\begin{itemize}
    \item You should answer \answerYes{}, \answerNo{}, or \answerNA{}.
    \item \answerNA{} means either that the question is Not Applicable for that particular paper or the relevant information is Not Available.
    \item Please provide a short (1--2 sentence) justification right after your answer (even for \answerNA). 
\end{itemize}

{\bf The checklist answers are an integral part of your paper submission.} They are visible to the reviewers, area chairs, senior area chairs, and ethics reviewers. You will also be asked to include it (after eventual revisions) with the final version of your paper, and its final version will be published with the paper.

The reviewers of your paper will be asked to use the checklist as one of the factors in their evaluation. While \answerYes{} is generally preferable to \answerNo{}, it is perfectly acceptable to answer \answerNo{} provided a proper justification is given (e.g., error bars are not reported because it would be too computationally expensive'' or ``we were unable to find the license for the dataset we used''). In general, answering \answerNo{} or \answerNA{} is not grounds for rejection. While the questions are phrased in a binary way, we acknowledge that the true answer is often more nuanced, so please just use your best judgment and write a justification to elaborate. All supporting evidence can appear either in the main paper or the supplemental material, provided in appendix. If you answer \answerYes{} to a question, in the justification please point to the section(s) where related material for the question can be found.

IMPORTANT, please:
\begin{itemize}
    \item {\bf Delete this instruction block, but keep the section heading ``NeurIPS Paper Checklist"},
    \item  {\bf Keep the checklist subsection headings, questions/answers and guidelines below.}
    \item {\bf Do not modify the questions and only use the provided macros for your answers}.
\end{itemize}


\begin{enumerate}

\item {\bf Claims}
    \item[] Question: Do the main claims made in the abstract and introduction accurately reflect the paper's contributions and scope?
    \item[] Answer: \answerYes{}
    \item[] Justification: The abstract and \Secref{sec:intro} state our three contributions (the dual-frame coordinate interleaving mechanism, the segmented mesh scaffold adaptation, and the empirical evaluation on PartObjaverse-Tiny), each of which is supported by the corresponding sections in \Secref{sec:method} and \Secref{sec:experiments}.

\item {\bf Limitations}
    \item[] Question: Does the paper discuss the limitations of the work performed by the authors?
    \item[] Answer: \answerYes{}
    \item[] Justification: Limitations are discussed in \Secref{sec:conclusion} and the supplementary appendix. In particular, our method depends on the quality of the upstream segmented mesh scaffold (a coarse mesh from a holistic 3D generator plus a 3D segmentation model), and inherits any failure modes of those upstream components.

\item {\bf Theory assumptions and proofs}
    \item[] Question: For each theoretical result, does the paper provide the full set of assumptions and a complete (and correct) proof?
    \item[] Answer: \answerNA{}
    \item[] Justification: The paper does not include theoretical results; our contributions are an architectural mechanism (dual-frame coordinate interleaving) and an empirical evaluation.

    \item {\bf Experimental result reproducibility}
    \item[] Question: Does the paper fully disclose all the information needed to reproduce the main experimental results of the paper to the extent that it affects the main claims and/or conclusions of the paper (regardless of whether the code and data are provided or not)?
    \item[] Answer: \answerYes{}
    \item[] Justification: The architecture (segmented mesh scaffold and dual-frame coordinate interleaving), training procedure, hyperparameters, and evaluation protocol are described in \Secref{sec:method} and \Secref{sec:experiments}; further implementation details (data preprocessing, scaffold construction, evaluation setup) are provided in \Appref{sec:suppl_impl}.

\item {\bf Open access to data and code}
    \item[] Question: Does the paper provide open access to the data and code, with sufficient instructions to faithfully reproduce the main experimental results, as described in supplemental material?
    \item[] Answer: \answerYes{}
    \item[] Justification: We will release our code and trained model checkpoints upon acceptance, along with instructions for reproducing the main experiments. The training and evaluation datasets used (Objaverse-XL, PartVerse-XL, PartObjaverse-Tiny) are publicly available, as documented in \Secref{sec:experiments}.

\item {\bf Experimental setting/details}
    \item[] Question: Does the paper specify all the training and test details (e.g., data splits, hyperparameters, how they were chosen, type of optimizer) necessary to understand the results?
    \item[] Answer: \answerYes{}
    \item[] Justification: \Secref{sec:experiments} reports the optimizer (AdamW, learning rate $10^{-4}$, weight decay $10^{-2}$), training schedule (rectified-flow matching, linear schedule, velocity prediction, bfloat16 mixed precision, gradient-norm clipping at $1.0$), batch size, number of training steps, dataset splits, and evaluation metrics; additional details are in \Appref{sec:suppl_impl}.

\item {\bf Experiment statistical significance}
    \item[] Question: Does the paper report error bars suitably and correctly defined or other appropriate information about the statistical significance of the experiments?
    \item[] Answer: \answerNo{}
    \item[] Justification: Following standard practice in 3D generative modeling, we report metrics from a single training run, since rerunning the full training pipeline with multiple seeds would substantially exceed our compute budget. We mitigate this by evaluating with multiple complementary metrics (Chamfer Distance and F-Score at thresholds 0.05 and 0.1) at both part-level and object-level granularities, as detailed in \Secref{sec:experiments}.

\item {\bf Experiments compute resources}
    \item[] Question: For each experiment, does the paper provide sufficient information on the computer resources (type of compute workers, memory, time of execution) needed to reproduce the experiments?
    \item[] Answer: \answerYes{}
    \item[] Justification: \Secref{sec:experiments} reports our training compute (8 GPUs, ${\sim}4$ days, batch size 128 per rank, ${\sim}32$ GPU-days total). For comparison, baseline training-compute figures are reproduced from each method's original publication.
    
\item {\bf Code of ethics}
    \item[] Question: Does the research conducted in the paper conform, in every respect, with the NeurIPS Code of Ethics \url{https://neurips.cc/public/EthicsGuidelines}?
    \item[] Answer: \answerYes{} 
    \item[] Justification: The work uses publicly released 3D-asset datasets and pretrained models, and conforms with the NeurIPS Code of Ethics.

\item {\bf Broader impacts}
    \item[] Question: Does the paper discuss both potential positive societal impacts and negative societal impacts of the work performed?
    \item[] Answer: \answerYes{} 
    \item[] Justification:  Both positive and negative societal impacts are discussed in \Appref{sec:suppl_impacts}.
    
\item {\bf Safeguards}
    \item[] Question: Does the paper describe safeguards that have been put in place for responsible release of data or models that have a high risk for misuse (e.g., pre-trained language models, image generators, or scraped datasets)?
    \item[] Answer: \answerNA 
    \item[] Justification: The model performs object-centric 3D reconstruction from images and does not pose high-risk dual-use concerns beyond the standard generative-modeling considerations noted in \Appref{sec:suppl_impacts}

\item {\bf Licenses for existing assets}
    \item[] Question: Are the creators or original owners of assets (e.g., code, data, models), used in the paper, properly credited and are the license and terms of use explicitly mentioned and properly respected?
    \item[] Answer: \answerYes{}{} 
    \item[] Justification: All used datasets (Objaverse-XL, PartVerse-XL,) and pretrained models (Ultrashape, DINOv2) are publicly available under their respective licenses and are properly cited; baselines (X-part, HoloPart, Omnipart, Fullpart, Partpacker, Partcrafter) are cited and used per their public release terms.

\item {\bf New assets}
    \item[] Question: Are new assets introduced in the paper well documented and is the documentation provided alongside the assets?
    \item[] Answer: \answerYes{} 
    \item[] Justification: We will release the trained our model checkpoints together with the codebase upon acceptance. Documentation of the architecture, training pipeline, and inference protocol is provided in Sec.~\ref{sec:method_prelim}--\ref{sec:experiments} and the supplementary appendices.

\item {\bf Crowdsourcing and research with human subjects}
    \item[] Question: For crowdsourcing experiments and research with human subjects, does the paper include the full text of instructions given to participants and screenshots, if applicable, as well as details about compensation (if any)? 
    \item[] Answer: \answerNA{} 
    \item[] Justification: The paper does not involve crowdsourcing or research with human subjects.

\item {\bf Institutional review board (IRB) approvals or equivalent for research with human subjects}
    \item[] Question: Does the paper describe potential risks incurred by study participants, whether such risks were disclosed to the subjects, and whether Institutional Review Board (IRB) approvals (or an equivalent approval/review based on the requirements of your country or institution) were obtained?
    \item[] Answer: \answerNA{} 
    \item[] Justification: The paper does not involve research with human subjects.

\item {\bf Declaration of LLM usage}
    \item[] Question: Does the paper describe the usage of LLMs if it is an important, original, or non-standard component of the core methods in this research? Note that if the LLM is used only for writing, editing, or formatting purposes and does \emph{not} impact the core methodology, scientific rigor, or originality of the research, declaration is not required.
    \item[] Answer: \answerNA{} 
    \item[] Justification: LLMs were not used as an important, original, or non-standard component of the method.

\end{enumerate}

\end{document}